\documentclass[times,review,10pt]{elsarticle}

\usepackage{amsmath}
\usepackage{amssymb}
\usepackage{graphicx}  
\usepackage{multirow}
\usepackage{booktabs}
\usepackage{makecell}
\usepackage{siunitx}
\usepackage{float}
\usepackage{url}
\usepackage{placeins}
\usepackage{lineno}

\journal{Pattern Recognition}
\begin{document}

\begin{frontmatter}

\title{Uncertainty-Guided Conservative Propagation for Structured Inference in Vessel Segmentation}
 
\affiliation[ksu]{
organization={Department of Computer Science, Kennesaw State University},
city={Marietta},
postcode={30060},
state={GA},
country={USA}
}

\affiliation[musc]{
organization={Department of Cardiology, Medical University of South Carolina},
city={Charleston},
state={SC},
country={USA}
}

\author[ksu]{Huan Huang}
\ead{hhuang10@students.kennesaw.edu}

\author[musc]{Michele Esposito}
\ead{espositm@musc.edu}

\author[ksu]{Chen Zhao\corref{cor1}}
\ead{czhao4@kennesaw.edu}

\cortext[cor1]{Corresponding author.}

%% Abstract
\begin{abstract}
Accurate and structurally consistent vessel segmentation is essential for downstream tasks such as vascular morphology analysis, stenosis assessment, and hemodynamic modeling. However, most existing methods directly use the single-pass output of a backbone network as the final prediction and lack an explicit model-internal mechanism that leverages predictive uncertainty and local structural information to further correct weak responses, vessel discontinuities, and structural inconsistencies. To address this limitation, we propose Uncertainty-Guided Conservative Propagation (UGCP), which formulates the initial segmentation logits as a finite-step state-update process. UGCP uses relative uncertainty between neighboring locations to regulate propagation direction, employs structure-aware modulation to suppress incompatible interactions, and introduces a source term to limit update drift. UGCP is differentiable and can be integrated into different segmentation networks for end-to-end optimization. We evaluated UGCP with five-fold cross-validation on four public datasets covering both 2D and 3D vessel segmentation, using generic convolutional networks, Transformers, and vessel-specific architectures. The results show that UGCP generally preserves regional overlap performance while improving centerline Dice (clDice) by $0.0032$--$0.0253$ and reducing Betti-0 error by approximately $10.9\%$--$60.1\%$ across all evaluated configurations. Ablation and visualization analyses further support the contributions of the proposed update components. These findings suggest the potential of UGCP to provide a more reliable segmentation basis for downstream analyses that depend on complete vascular geometry. The code is available at \url{https://github.com/chenzhao2023/UGCP_PR}.
\end{abstract}

%% Keywords
\begin{keyword}
Vessel Segmentation \sep Structured Inference \sep Uncertainty Quantification \sep Medical Image Segmentation \sep Deep Learning
\end{keyword}
\end{frontmatter}

% \linenumbers
%%%%%%%%%%%%%%%%%%%%%%%%%%%%%%%%%%%%%%%%%%%%%%%%%%%%%%%%%%%%%%%%%%
%%                         1. INTRODUCTION                       %%
%%%%%%%%%%%%%%%%%%%%%%%%%%%%%%%%%%%%%%%%%%%%%%%%%%%%%%%%%%%%%%%%%%
\section{Introduction}
 
Vessel segmentation is a critical task in medical image analysis, playing an important role in disease assessment and computer-aided diagnosis \cite{intro_cad_review}. It serves as a prerequisite for various downstream analyses, including centerline extraction, branch modeling, and stenosis evaluation \cite{intro_vessel_review}. Because these tasks critically depend on the structural integrity of vessels, segmentation quality is determined not only by local classification accuracy but also by the preservation of spatial continuity and topological consistency. Unlike conventional organ or lesion segmentation, vessel structures are thin, elongated, and highly branching \cite{intro_structure_review}, imposing strong spatial and topological dependencies. In practical imaging scenarios, noise, contrast variations, vessel overlap, and complex vascular patterns often lead to insufficient local evidence and increased prediction uncertainty \cite{intro_uq_review}. Under such conditions, local predictions may become unreliable, resulting in broken branches, discontinuities, or spurious connections. Therefore, achieving accurate and structurally consistent vessel segmentation remains challenging and important for downstream vascular analysis.

Existing methods address these challenges from four perspectives: enhancing feature representation, incorporating structural and topological knowledge, refining initial predictions, and modeling uncertainty. Convolutional neural networks (CNNs) improve vascular representation through multi-scale feature extraction and hierarchical feature modeling, while Transformer-based architectures further enhance long-range dependency and global contextual modeling \cite{intro_cnn_transformer_review}. Structure- and topology-aware methods improve morphological plausibility through connectivity and topological constraints, centerline supervision, or anatomical priors \cite{intro_topology_review, intro_topology_review2}. Post-processing methods refine initial predictions to suppress local noise and boundary errors \cite{intro_postprocessing_review}, whereas iterative inference methods further enable repeated prediction refinement \cite{intro_iterative_review}. Uncertainty modeling is used to identify ambiguous or unreliable regions and support uncertainty-aware supervision or learning \cite{intro_uq_method}.

However, existing approaches typically address these aspects in isolation, without a unified model-internal refinement mechanism. Architecture design and structural constraints primarily influence feature representation and initial prediction, while post-processing and iterative inference refine predictions without explicitly considering spatial variations in predictive reliability. Although uncertainty estimation provides information about prediction confidence, its integration into structure-preserving iterative refinement remains limited. Consequently, a unified refinement framework that simultaneously accounts for predictive uncertainty, vascular structure, and update stability remains underexplored.

To address this problem, we propose Uncertainty-Guided Conservative Propagation (UGCP), a differentiable, model-integrated, and end-to-end trainable module that refines backbone logits through a finite number of conservation-inspired local updates. The core idea is to allow relatively reliable low-uncertainty responses to support neighboring high-uncertainty regions. During updating, predictive uncertainty regulates the direction of local information flow, structure-aware modulation constrains interactions between neighboring regions, and a source term anchors the evolving state to the initial prediction to limit update drift. This coordinated update mechanism enables UGCP to selectively refine uncertain and structurally inconsistent regions while preserving relatively stable predictions. A preliminary version of UGCP was presented for coronary artery segmentation in our MICCAI study \cite{huan_miccai}; the present work substantially extends and generalizes the framework to broader vessel segmentation scenarios across different vascular anatomies, imaging modalities, data dimensionalities, and backbone architectures. The main contributions are summarized as follows:
\begin{itemize}
\item We formulate the initial vessel segmentation prediction as a finite-step, model-internal logit-state update process inspired by conservative interaction, enabling the backbone output to evolve through structured, end-to-end optimized refinement before the final prediction is produced.

\item We develop an uncertainty-guided local prediction update mechanism in which predictive reliability modulates the refinement behavior of different regions, structure-aware modulation suppresses erroneous interactions between incompatible regions, and a source term limits state drift during multi-step updates.

\item We design UGCP as a differentiable, backbone-agnostic module applicable to both 2D and 3D tasks, and systematically evaluate its effectiveness and applicability using convolutional networks, Transformers, and vessel-specific segmentation architectures across multiple datasets and imaging modalities.
\end{itemize}

%%%%%%%%%%%%%%%%%%%%%%%%%%%%%%%%%%%%%%%%%%%%%%%%%%%%%%%%%%%%%%%%%%
%%                         2. Related work                       %%
%%%%%%%%%%%%%%%%%%%%%%%%%%%%%%%%%%%%%%%%%%%%%%%%%%%%%%%%%%%%%%%%%%
\section{Related Work}

Existing vessel segmentation studies broadly fall into four directions: 1) general-purpose and vessel-specific architectures for improved feature modeling; 2) structural and topological constraints for morphological consistency; 3) post-processing or iterative inference for prediction refinement; and 4) uncertainty modeling for reliability assessment in ambiguous regions. These directions differ substantially in when and how structural information and predictive uncertainty are utilized.

\subsection{Vessel Segmentation Architectures}

General-purpose architectures such as U-Net and Transformer-based models provide strong backbones for vessel segmentation through multi-scale and contextual feature modeling \cite{rw_unet,rw_swinunetr}. Vessel-specific architectures further incorporate inductive biases for thin and tubular structures. FR-UNet uses full-resolution feature fusion, while DSCNet introduces geometry-adaptive dynamic snake convolutions \cite{rw_frunet,rw_dscnet}. Recent methods continue this direction: RetiNeXt combines frequency-domain enhancement with lightweight global--local feature modeling \cite{rw_retinext}, whereas Multi-scale Vision Mamba-UNet employs state-space modeling for long-range vascular representation \cite{rw_mvm_unet}.

These methods substantially improve vascular representation. Nevertheless, their main advances lie in the representation and mapping process from the input image to the segmentation output, and the final prediction is commonly produced through a single forward pass. When local image evidence is insufficient or ambiguous, stronger feature representations improve the quality of the initial prediction but do not directly ensure correction of residual local errors. Consequently, branch discontinuities, unstable boundaries, and spurious connections remain possible in the final segmentation.

\subsection{Structure- and Topology-Aware Learning}

Structure- and topology-aware methods incorporate vascular information into training objectives, explicit structural representations, or anatomical priors to improve continuity and morphological plausibility \cite{intro_structure_review,intro_topology_review}. One group of methods incorporates structural information into the training objective through topological, connectivity, or centerline-based constraints \cite{rw_topology_learning,rw_centerline_supervision}. These approaches encourage the model to preserve connected components, topological relationships, and tubular continuity in addition to pixel-wise overlap. Anatomical or morphological priors can further constrain the prediction space toward structurally plausible solutions \cite{rw_structural_prior}.

Another group of methods explicitly represents vascular structures rather than introducing them only through training constraints. Centerline-based approaches construct deformable representations of tubular structures \cite{rw_deformable_centerline}, whereas graph-based methods represent vascular branches and their topological relationships \cite{rw_graph_vessel}. DeepAtlas, a joint registration and segmentation framework, further exploits anatomical correspondences across images as auxiliary supervision to improve structural consistency \cite{rw_deepatlas}.

Collectively, these methods demonstrate the importance of structural information for vascular continuity and topological plausibility. However, structural knowledge is primarily incorporated through loss functions, auxiliary representations, or anatomical priors to shape the learned prediction function, while its use for adaptive correction of localized errors in an existing prediction remains relatively limited.

\subsection{Post-processing and Iterative Inference}

Beyond incorporating structural constraints during training, another line of research refines initial segmentation results through post-processing or iterative inference. Conventional post-processing methods include morphological operations and connected-component filtering \cite{intro_postprocessing_review}, vessel-specific refinement rules \cite{rw_vessel_postprocessing}, and conditional random fields (CRFs) \cite{rw_crf}. CRF-based methods impose pairwise compatibility relationships to promote local spatial consistency, whereas vessel-specific post-processing methods employ morphological or structural rules to improve connectivity and label consistency. Despite their effectiveness in suppressing spurious responses and resolving localized inconsistencies, these refinement procedures are generally separated from the backbone network and do not participate in end-to-end representation learning.

Iterative refinement has also been incorporated directly into neural networks. CRF-RNN unfolds mean-field inference in a conditional random field as recurrent neural network layers, enabling structured message passing to be jointly optimized with the preceding network \cite{rw_crf_rnn}. Partial differential equation (PDE)-inspired \cite{rw_pde} and diffusion-inspired approaches \cite{rw_diffusion} formulate optimization or dynamic evolution as a finite sequence of updates, thereby iteratively refining the initial prediction. These formulations extend segmentation beyond single-pass inference by progressively refining intermediate predictions through structured state updates.

Nevertheless, iterative methods differ considerably in their update criteria and propagation mechanisms. CRF and CRF-RNN approaches derive message passing from spatial proximity, appearance similarity, and pairwise compatibility \cite{rw_crf,rw_crf_rnn}, whereas diffusion- and PDE-inspired methods update predictions according to predefined or learnable smoothing terms, gradient terms, or energy-descent directions \cite{rw_pde,rw_diffusion}. In these formulations, predictive uncertainty is generally not an explicit component of the update rule, making propagation largely insensitive to spatial variations in prediction reliability.

\subsection{Uncertainty Modeling and Uncertainty-Guided Learning}

Uncertainty modeling aims to quantify prediction reliability and identify regions affected by insufficient image evidence, ambiguous class boundaries, or unstable model decisions \cite{intro_uq_review}. Common approaches include uncertainty estimation based on predictive probabilities or entropy, Monte Carlo dropout, deep ensembles, and evidential learning \cite{rw_uq}. Evidential learning models class-specific evidence and its corresponding Dirichlet distribution, providing an explicit description of evidence strength and reliability beyond normalized prediction probabilities \cite{rw_uq_evidential}. Existing uncertainty-aware segmentation methods predominantly employ uncertainty for error identification, reliability assessment, and quality control. However, uncertainty quantification is typically decoupled from the prediction update mechanism, with limited integration of uncertainty as an explicit signal for local segmentation refinement.

Uncertainty has also been incorporated into model training and supervision strategies. Uncertainty-guided segmentation methods adjust supervision according to prediction reliability \cite{rw_uq_guided_seg}, while uncertainty-aware semi-supervised learning uses uncertainty for pseudo-label selection, weighting, or supervision control \cite{rw_uq_semisupervised}. Uncertainty Rectified Pyramid Consistency (URPC) further incorporates uncertainty into consistency learning among multi-scale predictions, adaptively reducing the influence of unreliable predictions \cite{rw_urpc}.

Reliability-aware consistency, however, is distinct from explicit structural regularization. Existing uncertainty-guided strategies primarily modulate supervision or prediction consistency according to reliability, without directly constraining local vascular morphology or topology during refinement. The joint use of predictive uncertainty and local structural information within an iterative prediction-update mechanism therefore remains relatively underexplored.

In summary, existing studies have advanced vessel segmentation through improved feature representation, structural regularization, iterative refinement, and uncertainty modeling. However, the integration of predictive uncertainty and local structural constraints within a unified model-internal iterative refinement process remains limited.

%%%%%%%%%%%%%%%%%%%%%%%%%%%%%%%%%%%%%%%%%%%%%%%%%%%%%%%%%%%%%%%%%%
%%                         3. Methods                       %%
%%%%%%%%%%%%%%%%%%%%%%%%%%%%%%%%%%%%%%%%%%%%%%%%%%%%%%%%%%%%%%%%%%

\section{Methods}
\begin{figure}[!t]
\centering
\includegraphics[width=\linewidth]{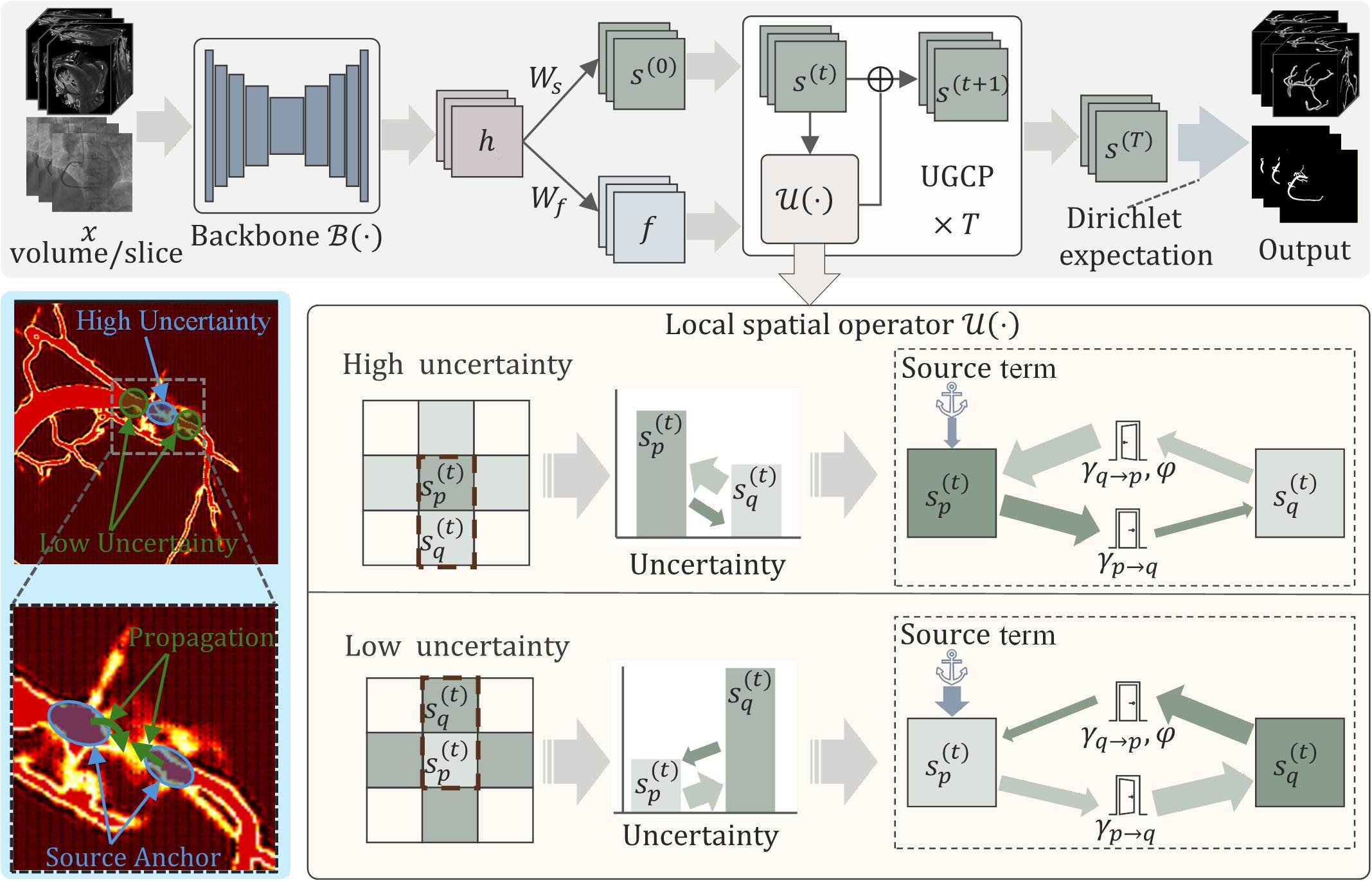}
 
\caption{
Overview of the proposed UGCP framework.
The top panel shows the overall pipeline, where the backbone output is mapped to the initial logit state and updated by UGCP for $T$ steps before generating the final prediction.
The bottom-left panel illustrates the core intuition of UGCP, where relatively reliable low-uncertainty responses guide the correction of neighboring high-uncertainty regions.
The bottom-right panel shows the local update operator $U(\cdot)$, where uncertainty-guided directional gating, structure-aware modulation, and source anchoring jointly regulate the state update; arrow thickness indicates the relative strength of directional information flow.
}
\label{fig:method}
\end{figure}
As shown in Fig.~\ref{fig:method}, we propose Uncertainty-Guided Conservative Propagation (UGCP), a differentiable module for structured refinement of vessel segmentation predictions. UGCP is inserted after the segmentation logit head and optimized end-to-end together with the backbone network. During inference, UGCP performs a finite-step update of the predicted logit state as part of the trained model, rather than being applied as a detached post-processing procedure after training.

%%%%%%%%%%%%%%%%%%%%%%%%%%%%%%%%%%%%%%%%%%%%%%%%%%%%%%%%%%%%%%%%%%
%%                         3.1 Problem Formulation            %%
%%%%%%%%%%%%%%%%%%%%%%%%%%%%%%%%%%%%%%%%%%%%%%%%%%%%%%%%%%%%%%%%%%
\subsection{Problem Formulation}
 
Given an input image $x$ defined on a discrete spatial domain $\Omega \subset \mathbb{Z}^d$, where $d \in \{2,3\}$ denotes the spatial dimension, let $y = \{ y_p \mid p \in \Omega \}$ denote the corresponding ground-truth segmentation, where $y_p \in \{0,1\}$ is the label at location $p$. The goal is to predict a segmentation $\hat{y}$ from $x$.

In conventional deep learning-based segmentation methods, predictions are typically generated via a single forward pass. Specifically, $x$ is first processed by a backbone network $B(\cdot)$ to extract feature representations $h = B(x)$, followed by a linear projection or classification head $W_s(\cdot)$ that maps the features to the logit state $s = W_s(h)$. The final class probabilities are then obtained via softmax. In this paradigm, predictions at each spatial location are produced through a direct mapping from features, without an explicit mechanism for iterative interaction or update during inference.

To address this limitation, we reformulate segmentation inference as a prediction update system in the logit space. Accordingly, the output of the segmentation head is treated as the initial state $s^{(0)} = W_s(h)$ and evolves through a finite number of structured updates before being mapped to probabilistic predictions, as shown in Eq.~(\ref{eq:update_state}).
\begin{equation}
s^{(t+1)} = s^{(t)} + \theta \, U\big(s^{(t)}\big), \quad t \in \{0,1,\dots,T-1\},
\label{eq:update_state}
\end{equation} where $s_p^{(t)} \in \mathbb{R}^{K}$ denotes the logit state at spatial location $p$ and update step $t$, and $K$ denotes the number of classes. The complete logit field is represented as $s^{(t)}=\{s_p^{(t)}\mid p\in\Omega\}$.
The parameter $\theta$ controls the update step size, $T$ denotes the number of update steps, and $U(\cdot)$ is a local update operator defined over spatial neighborhoods.

Under this formulation, conventional one-shot segmentation corresponds to the degenerate case with $T=0$, where no spatial interaction or prediction update is performed. In our formulation, spatial interaction is explicitly modeled through the update operator $U(\cdot)$, enabling progressive refinement of the initial prediction.

%%%%%%%%%%%%%%%%%%%%%%%%%%%%%%%%%%%%%%%%%%%%%%%%%%%%%%%%%%%%%%%%%%
%%   3.2 Conservation-Inspired Flux-Balance Formulation of Prediction Update   %%
%%%%%%%%%%%%%%%%%%%%%%%%%%%%%%%%%%%%%%%%%%%%%%%%%%%%%%%%%%%%%%%%%%
\subsection{Conservation-Inspired Flux-Balance Formulation of Prediction Update}
The key problem is to design a local update operator $U(\cdot)$ that governs how predictions are updated through spatial interaction. Independent updates modify each location using only local information, lacking explicit neighborhood coordination.  Aggregation-based approaches introduce spatial context through averaging, merging, or other neighborhood-based mixing operations, representing a simple form of local propagation \cite{method_aggre}. Despite introducing neighborhood information, these approaches remain value-level updates, where predictions are obtained via direct combination of neighboring responses without distinguishing the underlying update mechanisms. This limitation becomes more pronounced in iterative settings, where the output of each step is recursively fed into subsequent updates, increasing susceptibility to error accumulation and gradual structural degradation.

In contrast, we formulate prediction update as a state-change modeling problem rather than a value assignment process. This leads to a conservation-inspired flux-balance formulation with a source term. In the continuous setting, the evolution of the state variable is written in the partial differential equation (PDE) form in Eq.~(\ref{eq:conservative_pde}).

\begin{equation}
\frac{\partial s}{\partial t}
=
-\nabla \cdot F(s) + R(s),
\label{eq:conservative_pde}
\end{equation}
where $s$ denotes the state variable, $F(s)$ is the flux function, and $-\nabla \cdot F(s)$ represents the conservative flux-divergence term, interpreted as neural network information exchange between neighboring regions. The source term $R(s)$ accounts for non-conservative variations, representing update components that are not induced by neighborhood flux exchange.

Here, $s$ is instantiated as the logit representation, and the update is interpreted as a spatially structured process in the logit space. On a discrete spatial domain, the flux-divergence term $-\nabla\cdot F(s)$ reduces to the difference between incoming and outgoing fluxes. We define the axis-aligned local neighborhood as
$\mathcal{N}(p)=\{q\in\Omega \mid \|q-p\|_1=1\}$,
corresponding to a 4-neighborhood in 2D and a 6-neighborhood in 3D. Accordingly, the local update operator at location $p$ is given in Eq.~(\ref{eq:flux_balance}).

\begin{equation}
U(s_p^{(t)}) =
\sum_{q \in \mathcal{N}(p)}
\left(
F_{q \rightarrow p}^{(t)} - F_{p \rightarrow q}^{(t)}
\right)
+ R_p^{(t)},
\label{eq:flux_balance}
\end{equation}
where $F_{q \rightarrow p}^{(t)}$ and $F_{p \rightarrow q}^{(t)}$ denote the incoming and outgoing fluxes, respectively. The summation term is a discrete implementation of the flux-divergence term $-\nabla \cdot F(s)$ and is hereafter termed the flux-balance term, while $R_p^{(t)}$ denotes the source term at location $p$. The flux-balance term follows an in--out exchange structure and is conservation-inspired in the sense that state changes are modeled through directional exchanges between neighboring locations. Under this formulation, independent updates correspond to a degenerate case without explicit flux interaction, while simple averaging or merging corresponds to propagation with fixed or weakly selective flux.
%%%%%%%%%%%%%%%%%%%%%%%%%%%%%%%%%%%%%%%%%%%%%%%%%%%%%%%%%%%%%%%%%%
%%                         3.3 Evidential Representation and Uncertainty Modeling                       %%
%%%%%%%%%%%%%%%%%%%%%%%%%%%%%%%%%%%%%%%%%%%%%%%%%%%%%%%%%%%%%%%%%%
\subsection{Evidential Representation and Uncertainty Modeling}
During iterative updates, predictions interact through spatial propagation. Quantifying their reliability is therefore important, as unreliable estimates otherwise introduce unreliable information into neighboring states and increase the risk of structural inconsistency. Prediction probabilities and predictive entropy provide useful measures of class ambiguity \cite{method_UQ_review}. In UGCP, the uncertainty signal is recomputed from the evolving logit state at each update step and reflects the strength of evidence supporting the current prediction. We therefore adopt an evidential learning framework that maps the logit state to a Dirichlet distribution \cite{method_evidential_1}, enabling the expected class probabilities and their associated evidence strength to be represented within the same formulation.

For each spatial location $p \in \Omega$ and class index $k \in \{1, \dots, K\}$, where $K=2$ for binary segmentation, the Dirichlet concentration parameter is defined as Eq.~(\ref{eq:alpha}).
\begin{equation}
\alpha_{p,k}^{(t)} = \mathrm{softplus}\big(s_{p,k}^{(t)}\big) + 1
\label{eq:alpha}
\end{equation}
Here, the softplus transformation produces non-negative evidence, and adding the unit prior yields the corresponding Dirichlet concentration parameter.
The expected class probability is defined as Eq.~(\ref{eq:pi}).
\begin{equation}
\pi_{p,k}^{(t)} = \frac{\alpha_{p,k}^{(t)}}{\sum_{j=1}^{K} \alpha_{p,j}^{(t)} + \epsilon},
\label{eq:pi}
\end{equation}
where $\epsilon > 0$ ensures numerical stability. The Dirichlet strength $\sum_{j=1}^{K} \alpha_{p,j}^{(t)}$, which increases with the accumulated evidence, defines a continuous uncertainty measure as shown in Eq.~(\ref{eq:uncertainty}).
\begin{equation}
u_p^{(t)} = \frac{K}{\sum_{j=1}^{K} \alpha_{p,j}^{(t)} + \epsilon},
\label{eq:uncertainty}
\end{equation}
where smaller Dirichlet strength corresponds to higher predictive uncertainty.

This uncertainty field serves as a local control signal for information propagation, providing directional guidance and magnitude modulation.

%%%%%%%%%%%%%%%%%%%%%%%%%%%%%%%%%%%%%%%%%%%%%%%%%%%%%%%%%%%%%%%%%%
%%        3.4 Uncertainty-Guided Flux and Source Design       %%
%%%%%%%%%%%%%%%%%%%%%%%%%%%%%%%%%%%%%%%%%%%%%%%%%%%%%%%%%%%%%%%%%%
\subsection{Uncertainty-Guided Flux and Source Design}
The flux-balance term in Eq.~(\ref{eq:flux_balance}) models information exchange between neighboring locations. Without directional control, such propagation is susceptible to diffusion-like smoothing, leading to oversmoothing and loss of structural consistency. We therefore define uncertainty-guided directional gates for adjacent locations $p$ and $q$ to enable adaptive and directed information propagation, as shown in Eq.~(\ref{eq:uq_gate}).

\begin{equation}
\gamma_{q \rightarrow p}^{(t)}
=
\sigma
\left(
\frac{
u_p^{(t)} - u_q^{(t)}
}{
\tau + \epsilon
}
\right),
\quad
\gamma_{p \rightarrow q}^{(t)}
=
\sigma
\left(
\frac{
u_q^{(t)} - u_p^{(t)}
}{
\tau + \epsilon
}
\right),
\label{eq:uq_gate}
\end{equation}
where $\sigma(\cdot)$ denotes the sigmoid activation function and $\tau>0$ controls the sensitivity of the directional gates to uncertainty differences. A smaller $\tau$ produces a sharper directional preference, whereas a larger $\tau$ yields smoother and more balanced interactions. The two directional gates satisfy
$
\gamma_{q\rightarrow p}^{(t)}+\gamma_{p\rightarrow q}^{(t)}=1,
$
providing complementary directional weights for each neighboring pair. When $u_p^{(t)}>u_q^{(t)}$, location $p$ is less reliable than $q$, resulting in a larger incoming gate $\gamma_{q\rightarrow p}^{(t)}$ and a smaller reverse gate $\gamma_{p\rightarrow q}^{(t)}$. Thus, predictive uncertainty introduces an asymmetric correction preference according to the relative reliability of adjacent states.

While uncertainty-guided gates control propagation direction, they do not enforce structural consistency across neighboring locations. Near structural boundaries, direct propagation is susceptible to cross-structure contamination. Thus, we introduce a structure-aware edge modulation factor to regulate propagation across incompatible regions. A feature projection head $W_f(\cdot)$ maps the backbone feature $h$ to a decision-aligned feature space $f = W_f(h)$, and the structure-aware edge modulation factor between adjacent locations $p$ and $q$ is defined in Eq.~(\ref{eq:edge_factor}).

\begin{equation}
\phi_{p,q}
=
\tanh
\left(
w^\top (f_p - f_q)
\right),
\label{eq:edge_factor}
\end{equation}
where $f_p,f_q\in\mathbb{R}^{C_f}$ denote the projected features at locations $p$ and $q$, respectively, $C_f$ denotes the dimensionality of the projected feature space, and $w\in\mathbb{R}^{C_f}$ is a learnable projection vector. The ordered difference $f_p-f_q$ represents a directional, decision-related discrepancy between adjacent locations, while the projection $w^\top(f_p-f_q)$ determines how this discrepancy modulates the incoming contribution. The hyperbolic tangent bounds the modulation within $[-1,1]$, preventing unbounded amplification while retaining signed control. Values near zero suppress discrepancy-driven incoming correction, whereas positive and negative values respectively reinforce or counteract the incoming contribution in the ordered $q \rightarrow p$ interaction. Under these definitions, the incoming and outgoing fluxes are defined in Eq.~(\ref{eq:in_out_flux}).

\begin{equation}
F_{q \rightarrow p}^{(t)}
=
\gamma_{q \rightarrow p}^{(t)} \, \phi_{p,q} \, s_q^{(t)},
\quad
F_{p \rightarrow q}^{(t)}
=
\gamma_{p \rightarrow q}^{(t)} \, s_p^{(t)}.
\label{eq:in_out_flux}
\end{equation}

Here, $\phi_{p,q}$ is applied to the incoming component because cross-structure contamination occurs when incompatible neighbor responses are aggregated into the state of location $p$. The outgoing flux is governed by the uncertainty-guided gate, so that the contribution of $s_p^{(t)}$ remains reliability-driven rather than being additionally suppressed by local boundary discrepancies. Accordingly, the flux-balance term in Eq.~(\ref{eq:flux_balance}) is instantiated as in Eq.~(\ref{eq:neighborhood_interaction}).

\begin{equation}
\sum_{q \in \mathcal{N}(p)}
\left(
F_{q \rightarrow p}^{(t)} - F_{p \rightarrow q}^{(t)}
\right)
=
\sum_{q \in \mathcal{N}(p)}
\left(
\gamma_{q \rightarrow p}^{(t)} \phi_{p,q} s_q^{(t)}
-
\gamma_{p \rightarrow q}^{(t)} s_p^{(t)}
\right).
\label{eq:neighborhood_interaction}
\end{equation}

The flux-balance term characterizes conservation-inspired neighborhood interaction, while non-conservative variations arising during iterative updates are handled by an additional source term for stabilization. To control the strength of this source term, we define an uncertainty-dependent source gate as shown in Eq.~(\ref{eq:source_gate}).

\begin{equation}
r_p^{(t)}
=
\sigma
\left(
\frac{
u_0 - u_p^{(t)}
}{
\tau + \epsilon
}
\right),
\label{eq:source_gate}
\end{equation}
where $u_0$ is a reference uncertainty level that controls the relative strength of source anchoring. When $u_p^{(t)}<u_0$, the source gate satisfies $r_p^{(t)}>0.5$ and more strongly anchors the evolving state to the initial prediction. Conversely, when $u_p^{(t)}>u_0$, the anchoring strength is reduced, allowing the uncertain state greater flexibility to receive neighborhood-based correction. The source term is then formulated as Eq.~(\ref{eq:source_term}).

\begin{equation}
R_p^{(t)}
=
r_p^{(t)}
\left(
s_p^{(0)} - s_p^{(t)}
\right),
\label{eq:source_term}
\end{equation}
where $s_p^{(0)}$ denotes the initial logit state and $s_p^{(t)}$ denotes the current logit state at step $t$. This uncertainty-dependent source term protects relatively reliable states from unnecessary drift while preserving greater correction flexibility in uncertain regions.

Combining the uncertainty-guided directional gates, the structure-aware edge modulation factor, and the source term, the final local update operator is defined as Eq.~(\ref{eq:ugcp_operator}).

\begin{equation}
U(s_p^{(t)})
=
\sum_{q \in \mathcal{N}(p)}
\left(
\gamma_{q \rightarrow p}^{(t)}
\phi_{p,q}
s_q^{(t)}
-
\gamma_{p \rightarrow q}^{(t)}
s_p^{(t)}
\right)
+
r_p^{(t)}
\left(
s_p^{(0)} - s_p^{(t)}
\right).
\label{eq:ugcp_operator}
\end{equation}
The logit state is then updated by
\begin{equation}
s_p^{(t+1)}
=
s_p^{(t)}
+
\theta U(s_p^{(t)}),
\quad
t = 0,1,\dots,T-1.
\label{eq:pointwise_update}
\end{equation}

The accumulation of unreliable updates is jointly mitigated by the complementary uncertainty-guided gates, the suppression of structurally incompatible interactions, the uncertainty-dependent source anchoring, and the finite update process controlled by the number of steps $T$ and the step size $\theta$. After $T$ update steps, the final logit state $s^{(T)}$ is mapped to Dirichlet parameters, and the final class probabilities are obtained from the Dirichlet expectation.

Overall, UGCP reformulates segmentation inference as a structured logit-space update process, in which predictions are iteratively coordinated through uncertainty-guided propagation, structure-aware interaction, and source-based stabilization.

%%%%%%%%%%%%%%%%%%%%%%%%%%%%%%%%%%%%%%%%%%%%%%%%%%%%%%%%%%%%%%%%%%
%%        3.5 Implementation and Optimization       %%
%%%%%%%%%%%%%%%%%%%%%%%%%%%%%%%%%%%%%%%%%%%%%%%%%%%%%%%%%%%%%%%%%%
\subsection{Implementation and Optimization}

UGCP is implemented as a backbone-agnostic differentiable update module inserted after the segmentation logit head. It is optimized end-to-end together with the backbone network, rather than being applied as a detached post-processing procedure after model training. For the baseline models, the backbone and classification head produce logits, which are directly converted to the foreground probability map. For the corresponding UGCP variants, the same backbone and classification head generate the initial logit state, which is then updated by the proposed UGCP operator through a fixed number of differentiable update steps.

The training objective consists of segmentation supervision and evidential uncertainty regularization. For the baseline models, the training objective combines Dice loss and binary cross-entropy (BCE) loss as shown in Eq.~(\ref{eq:base_loss}).

\begin{equation}
\mathcal{L}_{base}
=
\mathcal{L}_{Dice}
+
\mathcal{L}_{BCE}.
\label{eq:base_loss}
\end{equation}

For UGCP, the final logit state is mapped to a Dirichlet evidence representation, and the expected Dirichlet probability is used for segmentation supervision. 
The overall objective is defined as Eq.~(\ref{eq:ugcp_loss}).

\begin{equation}
\mathcal{L}_{UGCP}
=
\mathcal{L}_{Dice}
+
\mathcal{L}_{BCE}
+
\lambda_{uq}\mathcal{L}_{UQ},
\label{eq:ugcp_loss}
\end{equation}

where $\mathcal{L}_{BCE}$ denotes the binary cross-entropy loss computed from the expected Dirichlet foreground probability $\pi_{p,1}^{(T)}$ and the binary ground-truth label $y_p \in \{0,1\}$. The hyperparameter $\lambda_{uq}$ controls the strength of evidential uncertainty regularization. Following evidential learning \cite{method_evidential_1}, the uncertainty regularization is defined as the average of a per-location evidential loss $\mathcal{L}_{UQ}
=
\frac{1}{|\Omega|}
\sum_{p \in \Omega}
\ell_{UQ,p}$, where $|\Omega|$ is the number of spatial locations and $\ell_{UQ,p}$ denotes the evidential loss at location $p$:

\begin{equation}
\begin{aligned}
\ell_{UQ,p}
=
&\sum_{k=1}^{K}
y_{p,k}
\left[
\psi(S_p^{(T)})
-
\psi(\alpha_{p,k}^{(T)})
\right]
+
\sum_{k=1}^{K}
\left(
\tilde{\alpha}_{p,k}^{(T)} - 1
\right)
\left[
\psi(\tilde{\alpha}_{p,k}^{(T)})
-
\psi(\tilde{S}_p^{(T)})
\right].
\end{aligned}
\label{eq:uq_loss_point}
\end{equation}

\noindent where $\psi(\cdot)$ denotes the digamma function, $S_p^{(T)}=\sum_{k=1}^{K}\alpha_{p,k}^{(T)}$, and $\tilde{S}_p^{(T)}=\sum_{k=1}^{K}\tilde{\alpha}_{p,k}^{(T)}$. The adjusted concentration parameter is defined component-wise as
\begin{equation}
\tilde{\alpha}_{p,k}^{(T)}
=
y_{p,k}
+
\left(1-y_{p,k}\right)
\alpha_{p,k}^{(T)},
\label{eq:adjusted_alpha}
\end{equation}
where $y_{p,k}$ denotes the one-hot ground-truth label for class $k$. The first summation in Eq.~(\ref{eq:uq_loss_point}) corresponds to the evidential negative log-likelihood, encouraging evidence for the ground-truth class.  The second summation regularizes the adjusted concentration parameter and penalizes unsupported evidence assigned to incorrect classes, thereby reducing over-confident predictions when evidence is insufficient.  

%%%%%%%%%%%%%%%%%%%%%%%%%%%%%%%%%%%%%%%%%%%%%%%%%%%%%%%%%%%%%%%%%%
%%        4   Experiment and Results       %%
%%%%%%%%%%%%%%%%%%%%%%%%%%%%%%%%%%%%%%%%%%%%%%%%%%%%%%%%%%%%%%%%%%
\section{Experiments and Results}
%%%%%%%%%%%%%%%%%%%%%%%%%%%%%%%%%%%%%%%%%%%%%%%%%%%%%%%%%%%%%%%%%%
%%        4.1   Experimental Setup       %%
%%%%%%%%%%%%%%%%%%%%%%%%%%%%%%%%%%%%%%%%%%%%%%%%%%%%%%%%%%%%%%%%%%
\subsection{Experimental Setup}

\textbf{Datasets and preprocessing.} 
We evaluated UGCP on four public datasets covering multiple imaging modalities in both 2D and 3D settings: FIVES with 800 fundus images \cite{data_fives}, the invasive coronary angiography (ICA) dataset with 616 images \cite{data_ica}, ImageCAS with 1,000 coronary computed tomography angiography (CCTA) cases \cite{data_imagecas}, and COSTA with 355 time-of-flight magnetic resonance angiography (TOF-MRA) cases \cite{data_brain}.

For the 2D datasets, FIVES images have a resolution of $2048 \times 2048$ with three color channels and were converted to single-channel inputs and randomly cropped into $512 \times 512$ patches during training. ICA images were used at their original resolution of $512 \times 512$. Both datasets used random horizontal and vertical flipping, random rotation, intensity scaling, and Gaussian noise for data augmentation. During inference, FIVES was evaluated using a $512 \times 512$ sliding window with an overlap ratio of 0.5, whereas ICA was evaluated at full resolution.

For the 3D datasets, ImageCAS and COSTA volumes were resampled to an isotropic voxel spacing of $0.5 \times 0.5 \times 0.5$ mm. ImageCAS additionally used CCTA-specific windowing with a window level of 300 and a window width of 800. Patch sizes of $96 \times 96 \times 96$ and $64 \times 64 \times 64$ were used for ImageCAS and COSTA, respectively, with 16 patches randomly sampled per case during each epoch. Sliding-window inference used the corresponding patch sizes with an overlap ratio of 0.5.

All 2D images and 3D volumes were normalized to $[0,1]$. When official training and testing splits were provided, all available annotated data were merged and repartitioned using five-fold cross-validation. The partitions were generated at the image level for the 2D datasets and at the case level for the 3D datasets.

\textbf{Comparison methods.}
For the 2D experiments, we used 2D U-Net \cite{rw_unet}, 2D SwinUNETR \cite{rw_swinunetr}, and the vessel-specific FR-UNet \cite{rw_frunet}. For the 3D experiments, we used 3D U-Net \cite{rw_unet}, 3D SwinUNETR \cite{rw_swinunetr}, and the tubular-structure-specific DSCNet \cite{rw_dscnet}. Each baseline and its corresponding UGCP variant used the same data split and backbone-specific training settings, which were configured based on the original publication and official implementation. The baseline models and UGCP variants were optimized using the objectives in Eqs.~(\ref{eq:base_loss}) and (\ref{eq:ugcp_loss}), respectively.

\textbf{Implementation details.}
All models were implemented in PyTorch 2.4 and trained on an NVIDIA RTX 4090 GPU with 24 GB of memory. AdamW was used with a weight decay of $1 \times 10^{-4}$ and a cosine learning-rate schedule. All models were trained for a maximum of 800 epochs, with an early-stopping patience of 60 epochs based on the Dice similarity coefficient (DSC) on the validation fold. U-Net and SwinUNETR used an initial learning rate of $10^{-3}$ and batch sizes of 4 and 2 for the 2D and 3D settings, respectively. FR-UNet and DSCNet used an initial learning rate of $10^{-4}$ and batch sizes of 2 and 1, respectively. The uncertainty loss weight $\lambda_{uq}$ was set to 0.1 and 0.2 for the 2D and 3D experiments, respectively.

For UGCP, the number of update steps was set to $T=2$, the update step size to $\theta=1$, and the reference uncertainty level to $u_0=0.5$. The temperature parameter $\tau$ was set to $0.01$ for the 2D datasets and $0.1$ for the 3D datasets. These settings were fixed across all folds, and the sensitivity of $T$, $\theta$, $\tau$, and $u_0$ is further analyzed in Section~\ref{sec:ablation}.

\textbf{Evaluation metrics and statistical analysis.}
Segmentation performance was evaluated using DSC, centerline Dice (clDice), the 95th percentile Hausdorff distance (HD95), and Betti-0 error (B0E). For the 3D datasets, HD95 was reported in millimeters. Because FIVES and ICA do not provide reliable physical pixel spacing, HD95 for the 2D datasets was reported in pixels. B0E was defined as
$\left|\beta_0(\hat{y})-\beta_0(y)\right|$,
where $\hat{y}$ and $y$ denote the predicted and ground-truth binary masks, respectively, and $\beta_0(\cdot)$ denotes the number of connected components. Connected components were identified using 8-connectivity in 2D and 26-connectivity in 3D.

Five-fold results were reported as mean $\pm$ standard deviation across folds. Statistical analyses used all out-of-fold image-level or case-level results from the five folds. For each baseline and its corresponding UGCP variant, two-sided paired $t$-tests were used for DSC, clDice, and HD95, whereas a two-sided Wilcoxon signed-rank test was used for B0E. Statistical significance was set at $p<0.05$.

 %%%%%%%%%%%%%%%%%%%%%%%%%%%%%%%%%%%%%%%%%%%%%%%%%%%%%%%%%%%%%%%%%%
%%        4.2   Segmentation Results       %%
%%%%%%%%%%%%%%%%%%%%%%%%%%%%%%%%%%%%%%%%%%%%%%%%%%%%%%%%%%%%%%%%%%
\subsection{Segmentation Results}

\textbf{Quantitative results.}
The 3D results are presented in Table~\ref{tab:main_results_3d}. UGCP consistently improves clDice and reduces HD95 and B0E across all configurations. On COSTA with SwinUNETR, DSC and clDice increase by $0.0230$ and $0.0253$, respectively, while HD95 decreases from $5.94$ to $2.09$ mm and B0E from $105.90$ to $43.92$. On ImageCAS with DSCNet, DSC slightly decreases from $0.8193$ to $0.8190$, whereas clDice, HD95, and B0E all improve. These results indicate that UGCP improves structural consistency across different 3D backbones, including a tubular-structure-specific architecture.

\begin{table}[!t]
\centering
\small
\renewcommand{\arraystretch}{0.92}
\setlength{\tabcolsep}{2pt}
\setlength{\aboverulesep}{1pt}
\setlength{\belowrulesep}{1pt}
\caption{
Quantitative results on the 3D datasets.
Best results for each dataset and backbone are shown in bold.
Asterisks indicate statistically significant differences between each UGCP variant and its corresponding baseline:
$^{*}p<0.05$ and $^{**}p<0.01$.
HD95 is reported in millimeters.
}
\label{tab:main_results_3d}

\resizebox{\linewidth}{!}{%
\begin{tabular}{@{} l l l c c c c @{}}
\toprule
Dataset & Backbone & Method
& DSC $\uparrow$
& clDice $\uparrow$
& HD95 (mm) $\downarrow$
& B0E $\downarrow$ \\
\midrule

\multirow{6}{*}{COSTA}
& \multirow{2}{*}{U-Net}
& Baseline
& $.8422 \pm .0043$
& $.8878 \pm .0033$
& $5.07 \pm 0.86$
& $123.25 \pm 5.37$ \\

&
& +UGCP
& $\mathbf{.8542 \pm .0037}^{**}$
& $\mathbf{.9030 \pm .0044}^{**}$
& $\mathbf{3.34 \pm 0.53}^{**}$
& $\mathbf{67.60 \pm 4.35}^{**}$ \\

\cmidrule(lr){2-7}

& \multirow{2}{*}{SwinUNETR}
& Baseline
& $.8560 \pm .0025$
& $.8902 \pm .0058$
& $5.94 \pm 2.09$
& $105.90 \pm 9.62$ \\

&
& +UGCP
& $\mathbf{.8790 \pm .0028}^{**}$
& $\mathbf{.9155 \pm .0037}^{**}$
& $\mathbf{2.09 \pm 0.43}^{**}$
& $\mathbf{43.92 \pm 2.43}^{**}$ \\

\cmidrule(lr){2-7}

& \multirow{2}{*}{DSCNet}
& Baseline
& $.8682 \pm .0041$
& $.8968 \pm .0066$
& $5.26 \pm 2.26$
& $97.33 \pm 5.28$ \\

&
& +UGCP
& $\mathbf{.8815 \pm .0062}^{**}$
& $\mathbf{.9142 \pm .0047}^{**}$
& $\mathbf{2.39 \pm 0.55}^{**}$
& $\mathbf{38.84 \pm 4.84}^{**}$ \\

\midrule

\multirow{6}{*}{ImageCAS}
& \multirow{2}{*}{U-Net}
& Baseline
& $.8147 \pm .0051$
& $.8553 \pm .0038$
& $8.89 \pm 0.56$
& $49.37 \pm 8.25$ \\

&
& +UGCP
& $\mathbf{.8188 \pm .0040}^{**}$
& $\mathbf{.8585 \pm .0045}^{*}$
& $\mathbf{8.76 \pm 0.43}^{*}$
& $\mathbf{38.66 \pm 7.38}^{**}$ \\

\cmidrule(lr){2-7}

& \multirow{2}{*}{SwinUNETR}
& Baseline
& $.8160 \pm .0067$
& $.8513 \pm .0079$
& $8.83 \pm 0.60$
& $35.32 \pm 7.71$ \\

&
& +UGCP
& $\mathbf{.8189 \pm .0064}^{*}$
& $\mathbf{.8601 \pm .0035}^{**}$
& $\mathbf{8.56 \pm 0.42}^{*}$
& $\mathbf{30.13 \pm 7.19}^{**}$ \\

\cmidrule(lr){2-7}

& \multirow{2}{*}{DSCNet}
& Baseline
& $\mathbf{.8193 \pm .0088}$
& $.8610 \pm .0068$
& $7.97 \pm 0.72$
& $32.32 \pm 6.88$ \\

&
& +UGCP
& $.8190 \pm .0079$
& $\mathbf{.8656 \pm .0065}^{*}$
& $\mathbf{7.63 \pm 0.61}^{*}$
& $\mathbf{28.44 \pm 6.41}^{**}$ \\

\bottomrule
\end{tabular}%
}

\end{table}

\begin{table}[!t]
\centering
\small
\renewcommand{\arraystretch}{0.92}
\setlength{\aboverulesep}{1pt}
\setlength{\belowrulesep}{1pt}
\setlength{\tabcolsep}{2pt}
\caption{
Quantitative results on the 2D datasets.
Best results for each dataset and backbone are shown in bold.
Asterisks indicate statistically significant differences between each UGCP variant and its corresponding baseline:
$^{*}p<0.05$ and $^{**}p<0.01$.
HD95 is reported in pixels.
}
\label{tab:main_results_2d}

\resizebox{\linewidth}{!}{%
\begin{tabular}{@{} l l l c c c c @{}}
\toprule
Dataset & Backbone & Method
& DSC $\uparrow$
& clDice $\uparrow$
& HD95 $\downarrow$
& B0E $\downarrow$ \\
\midrule

\multirow{6}{*}{FIVES}
& \multirow{2}{*}{U-Net}
& Baseline
& $.8780 \pm .0089$
& $.8777 \pm .0110$
& $62.73 \pm 9.71$
& $120.90 \pm 10.50$ \\

&
& +UGCP
& $\mathbf{.8831 \pm .0090}^{**}$
& $\mathbf{.8833 \pm .0109}^{**}$
& $\mathbf{50.70 \pm 9.98}^{**}$
& $\mathbf{89.10 \pm 6.20}^{**}$ \\

\cmidrule(lr){2-7}

& \multirow{2}{*}{SwinUNETR}
& Baseline
& $.8668 \pm .0090$
& $.8614 \pm .0108$
& $68.48 \pm 12.49$
& $161.98 \pm 6.74$ \\

&
& +UGCP
& $\mathbf{.8711 \pm .0087}^{**}$
& $\mathbf{.8654 \pm .0103}^{*}$
& $\mathbf{58.08 \pm 11.48}^{**}$
& $\mathbf{130.66 \pm 10.50}^{**}$ \\

\cmidrule(lr){2-7}

& \multirow{2}{*}{FR-UNet}
& Baseline
& $.8776 \pm .0102$
& $.8841 \pm .0085$
& $58.43 \pm 9.13$
& $103.02 \pm 7.13$ \\

&
& +UGCP
& $\mathbf{.8793 \pm .0098}$
& $\mathbf{.8895 \pm .0092}^{**}$
& $\mathbf{47.44 \pm 10.12}^{**}$
& $\mathbf{85.21 \pm 6.65}^{**}$ \\

\midrule

\multirow{6}{*}{ICA}
& \multirow{2}{*}{U-Net}
& Baseline
& $.8932 \pm .0160$
& $.8979 \pm .0202$
& $27.41 \pm 4.91$
& $17.16 \pm 5.42$ \\

&
& +UGCP
& $\mathbf{.9045 \pm .0148}^{**}$
& $\mathbf{.9063 \pm .0198}^{**}$
& $\mathbf{24.76 \pm 4.43}^{**}$
& $\mathbf{15.29 \pm 3.57}^{**}$ \\

\cmidrule(lr){2-7}

& \multirow{2}{*}{SwinUNETR}
& Baseline
& $.8888 \pm .0104$
& $.8765 \pm .0119$
& $28.72 \pm 3.30$
& $24.16 \pm 5.89$ \\

&
& +UGCP
& $\mathbf{.8923 \pm .0088}^{**}$
& $\mathbf{.8838 \pm .0103}^{**}$
& $\mathbf{26.37 \pm 2.95}^{*}$
& $\mathbf{21.42 \pm 5.24}^{**}$ \\

\cmidrule(lr){2-7}

& \multirow{2}{*}{FR-UNet}
& Baseline
& $.8974 \pm .0118$
& $.9015 \pm .0106$
& $24.11 \pm 4.11$
& $16.05 \pm 5.32$ \\

&
& +UGCP
& $\mathbf{.8991 \pm .0095}^{*}$ 
& $\mathbf{.9135 \pm .0115}^{**}$
& $\mathbf{22.44 \pm 3.69}^{**}$
& $\mathbf{13.13 \pm 4.87}^{**}$ \\

\bottomrule
\end{tabular}%
}

\end{table}

The 2D results are presented in Table~\ref{tab:main_results_2d}. A similar trend is observed across the 2D datasets, with consistent improvements in clDice and reductions in HD95 and B0E across all backbones. On FIVES, HD95 decreases by $12.03$ and $10.40$ pixels for U-Net and SwinUNETR, respectively. On ICA with FR-UNet, DSC increases from $0.8974$ to $0.8991$ and clDice from $0.9015$ to $0.9135$, while HD95 and B0E decrease from $24.11$ to $22.44$ and from $16.05$ to $13.13$, respectively. These results further indicate that UGCP improves structural prediction quality when integrated with a vessel-specific backbone.

\begin{figure}[!t]
\centering
\includegraphics[width=\linewidth]{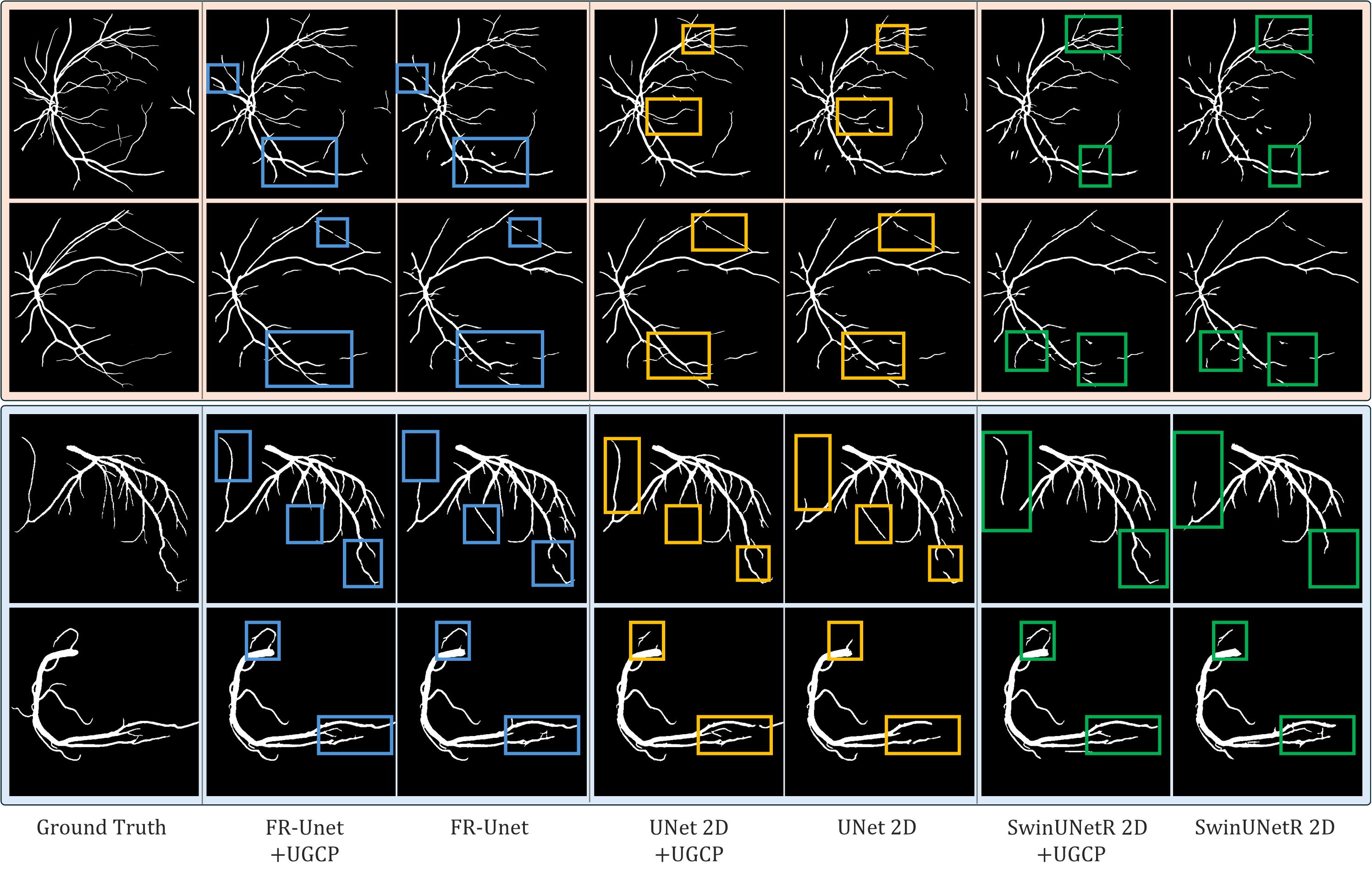}
 
% =========================
% 2D qualitative
% =========================
\caption{
Qualitative comparison between baseline and UGCP predictions on the 2D datasets.
Representative examples from FIVES are shown in rows 1--2 and those from ICA in rows 3--4.
}
\label{fig:qualitative_2d}
\end{figure}

\begin{figure}[t]
\centering
\includegraphics[width=\linewidth]{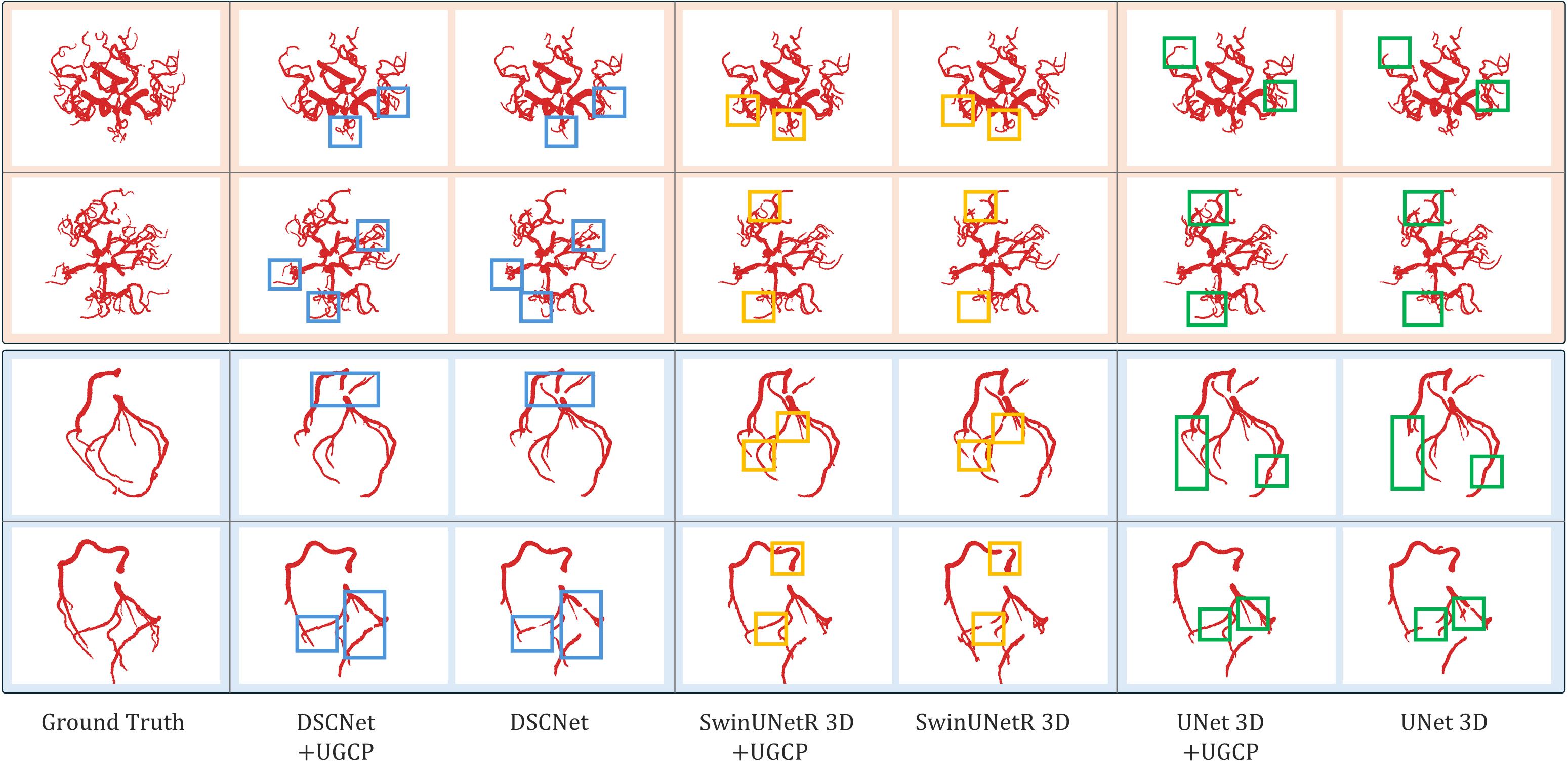}
% =========================
% 3D qualitative
% =========================
\caption{
Qualitative comparison between baseline and UGCP predictions on the 3D datasets.
Representative examples from COSTA are shown in rows 1--2 and those from ImageCAS in rows 3--4.
}
\label{fig:qualitative_3d}
\end{figure}

\textbf{Qualitative results.}
Figs.~\ref{fig:qualitative_2d} and \ref{fig:qualitative_3d} show representative comparisons on the four datasets. Compared with the baseline predictions, UGCP produces more continuous thin-vessel structures and fewer local discontinuities, particularly in the highlighted regions. In addition, isolated or spurious false-positive responses are reduced, indicating that the uncertainty-guided and structure-aware interactions not only support weak vessel responses but also suppress unreliable local predictions. These improvements are observed across both 2D and 3D datasets and are consistent with the quantitative gains in clDice, HD95, and B0E reported in Tables~\ref{tab:main_results_3d} and \ref{tab:main_results_2d}.

 %%%%%%%%%%%%%%%%%%%%%%%%%%%%%%%%%%%%%%%%%%%%%%%%%%%%%%%%%%%%%%%%%%
%%        4.3    Ablation and Sensitivity Analysis       %%
%%%%%%%%%%%%%%%%%%%%%%%%%%%%%%%%%%%%%%%%%%%%%%%%%%%%%%%%%%%%%%%%%%
\subsection{Ablation and Sensitivity Analysis}
\label{sec:ablation}
We conducted ablation studies on ICA and ImageCAS using U-Net and SwinUNETR to examine the effect of evidential uncertainty supervision, the contributions of the UGCP components, and the sensitivity to key hyperparameters.
 
\textbf{Component ablation.}
As shown in Table~\ref{tab:ablation_uq}, using the evidentially regularized objective $\mathcal{L}_{UGCP}$ without the UGCP update module does not yield consistent improvements over $\mathcal{L}_{base}$. In contrast, combining the same evidential objective with UGCP improves all metrics across both datasets and backbones. This indicates that the performance gains mainly arise from the uncertainty-guided structured updates rather than evidential supervision alone.

\begin{table}[!t]
\centering
\small
\renewcommand{\arraystretch}{0.92}
\setlength{\tabcolsep}{2pt}
\setlength{\aboverulesep}{1pt}
\setlength{\belowrulesep}{1pt} 
  
\caption{
Ablation of evidential uncertainty supervision and UGCP across different backbones.
$\mathcal{L}_{base}$ denotes the Dice and BCE objective in Eq.~(\ref{eq:base_loss}), whereas $\mathcal{L}_{UGCP}$ denotes the evidentially regularized objective in Eq.~(\ref{eq:ugcp_loss}).
Best results for each dataset and backbone are shown in bold.
HD95 is reported in pixels for ICA and in millimeters for ImageCAS.
}
\label{tab:ablation_uq}

\resizebox{\linewidth}{!}{%
\begin{tabular}{@{} l l c c c c c c c @{}}
\toprule
Dataset
& Backbone
& $\mathcal{L}_{base}$
& $\mathcal{L}_{UGCP}$
& UGCP
& DSC $\uparrow$
& clDice $\uparrow$
& HD95 $\downarrow$ 
& B0E $\downarrow$ \\
\midrule

\multirow{6}{*}{ICA}
& \multirow{3}{*}{U-Net}
& \checkmark
& --
& --
& $.8932 \pm .0160$
& $.8979 \pm .0202$
& $27.41 \pm 4.91$
& $17.16 \pm 5.42$ \\

&
& --
& \checkmark
& --
& $.8816 \pm .0084$
& $.8760 \pm .0082$
& $26.60 \pm 1.68$
& $17.55 \pm 5.11$ \\

&
& --
& \checkmark
& \checkmark
& $\mathbf{.9045 \pm .0148}$
& $\mathbf{.9063 \pm .0198}$
& $\mathbf{24.76 \pm 4.43}$
& $\mathbf{15.29 \pm 3.57}$ \\

\cmidrule(lr){2-9}

& \multirow{3}{*}{SwinUNETR}
& \checkmark
& --
& --
& $.8888 \pm .0104$
& $.8765 \pm .0119$
& $28.72 \pm 3.30$
& $24.16 \pm 5.89$ \\

&
& --
& \checkmark
& --
& $.8866 \pm .0092$
& $.8772 \pm .0144$
& $29.31 \pm 3.22$
& $30.13 \pm 5.77$ \\

&
& --
& \checkmark
& \checkmark
& $\mathbf{.8923 \pm .0088}$
& $\mathbf{.8838 \pm .0103}$
& $\mathbf{26.37 \pm 2.95}$
& $\mathbf{21.42 \pm 5.24}$ \\

\midrule

\multirow{6}{*}{ImageCAS}
& \multirow{3}{*}{U-Net}
& \checkmark
& --
& --
& $.8147 \pm .0051$
& $.8553 \pm .0038$
& $8.89 \pm 0.56$
& $49.37 \pm 8.25$ \\

&
& --
& \checkmark
& --
& $.8162 \pm .0088$
& $.8541 \pm .0053$
& $8.95 \pm 0.66$
& $49.39 \pm 8.98$ \\

&
& --
& \checkmark
& \checkmark
& $\mathbf{.8188 \pm .0040}$
& $\mathbf{.8585 \pm .0045}$
& $\mathbf{8.76 \pm 0.43}$
& $\mathbf{38.66 \pm 7.38}$ \\

\cmidrule(lr){2-9}

& \multirow{3}{*}{SwinUNETR}
& \checkmark
& --
& --
& $.8160 \pm .0067$
& $.8513 \pm .0079$
& $8.83 \pm 0.60$
& $35.32 \pm 7.71$ \\

&
& --
& \checkmark
& --
& $.8132 \pm .0076$
& $.8413 \pm .0072$
& $9.23 \pm 0.84$
& $35.73 \pm 8.18$ \\

&
& --
& \checkmark
& \checkmark
& $\mathbf{.8189 \pm .0064}$
& $\mathbf{.8601 \pm .0035}$
& $\mathbf{8.56 \pm 0.42}$
& $\mathbf{30.13 \pm 7.19}$ \\

\bottomrule
\end{tabular}%
}

\end{table}

\begin{table}[!t]
\centering
\small
\renewcommand{\arraystretch}{0.82}
\setlength{\extrarowheight}{0pt}
\setlength{\tabcolsep}{1.5pt}
\setlength{\aboverulesep}{0.5pt}
\setlength{\belowrulesep}{0.5pt}

\caption{
Component analysis of UGCP across different backbones.
$\gamma$, $\phi$, and $R$ denote the uncertainty-guided gate, structure-aware modulation factor, and source term, respectively.
All configurations are trained using $\mathcal{L}_{UGCP}$, and the first configuration represents evidential uncertainty supervision without the proposed update components.
Best results for each dataset and backbone are shown in bold.
HD95 is reported in pixels for ICA and in millimeters for ImageCAS.
}
\label{tab:component_analysis}

\begin{tabular*}{\linewidth}
{@{\extracolsep{\fill}} l l c c c c c c c @{}}
\toprule
Dataset
& Backbone
& $\gamma$
& $\phi$
& $R$
& DSC $\uparrow$
& clDice $\uparrow$
& HD95 $\downarrow$ 
& B0E $\downarrow$ \\
\midrule

\multirow{8}{*}{ICA}
& \multirow{4}{*}{U-Net}
& --
& --
& --
& $.8816 \pm .0084$
& $.8760 \pm .0082$
& $26.60 \pm 1.68$ 
& $17.55 \pm 5.11$ \\

&
& \checkmark
& --
& --
& $.8854 \pm .0086$
& $.8782 \pm .0087$
& $26.42 \pm 1.58$ 
& $17.02 \pm 4.14$ \\

&
& \checkmark
& \checkmark
& --
& $.8911 \pm .0095$
& $.8783 \pm .0064$
& $25.57 \pm 1.31$ 
& $16.64 \pm 4.55$ \\

&
& \checkmark
& \checkmark
& \checkmark
& $\mathbf{.9045 \pm .0148}$
& $\mathbf{.9063 \pm .0198}$
& $\mathbf{24.76 \pm 4.43}$
& $\mathbf{15.29 \pm 3.57}$ \\

\cmidrule(lr){2-9}

& \multirow{4}{*}{SwinUNETR}
& --
& --
& --
& $.8866 \pm .0092$
& $.8772 \pm .0144$
& $29.31 \pm 3.22$ 
& $30.13 \pm 5.77$ \\

&
& \checkmark
& --
& --
& $.8875 \pm .0121$
& $.8742 \pm .0092$
& $27.74 \pm 3.18$ 
& $24.33 \pm 4.79$ \\

&
& \checkmark
& \checkmark
& --
& $.8904 \pm .0085$
& $.8720 \pm .0111$
& $26.58 \pm 3.23$ 
& $24.08 \pm 5.06$ \\

&
& \checkmark
& \checkmark
& \checkmark
& $\mathbf{.8923 \pm .0088}$
& $\mathbf{.8838 \pm .0103}$
& $\mathbf{26.37 \pm 2.95}$
& $\mathbf{21.42 \pm 5.24}$ \\

\midrule

\multirow{8}{*}{ImageCAS}
& \multirow{4}{*}{U-Net}
& --
& --
& --
& $.8162 \pm .0088$
& $.8541 \pm .0053$
& $8.95 \pm 0.66$
& $49.39 \pm 8.98$ \\

&
& \checkmark
& --
& --
& $.8156 \pm .0079$
& $.8562 \pm .0044$
& $8.95 \pm 0.77$
& $47.13 \pm 5.95$ \\

&
& \checkmark
& \checkmark
& --
& $.8160 \pm .0081$
& $.8544 \pm .0051$
& $8.82 \pm 0.28$
& $41.28 \pm 7.74$ \\

&
& \checkmark
& \checkmark
& \checkmark
& $\mathbf{.8188 \pm .0040}$
& $\mathbf{.8585 \pm .0045}$
& $\mathbf{8.76 \pm 0.43}$
& $\mathbf{38.66 \pm 7.38}$ \\

\cmidrule(lr){2-9}

& \multirow{4}{*}{SwinUNETR}
& --
& --
& --
& $.8132 \pm .0076$
& $.8413 \pm .0072$
& $9.23 \pm 0.84$
& $35.73 \pm 8.18$ \\

&
& \checkmark
& --
& --
& $.8155 \pm .0082$
& $.8534 \pm .0071$
& $8.72 \pm 0.59$
& $35.08 \pm 6.54$ \\

&
& \checkmark
& \checkmark
& --
& $.8172 \pm .0108$
& $.8547 \pm .0114$
& $8.67 \pm 0.55$
& $33.15 \pm 8.22$ \\

&
& \checkmark
& \checkmark
& \checkmark
& $\mathbf{.8189 \pm .0064}$
& $\mathbf{.8601 \pm .0035}$
& $\mathbf{8.56 \pm 0.42}$
& $\mathbf{30.13 \pm 7.19}$ \\

\bottomrule
\end{tabular*}
\end{table}

Table~\ref{tab:component_analysis} further evaluates the cumulative contributions of $\gamma$, $\phi$, and $R$. The uncertainty-guided gate $\gamma$ generally improves structural and geometric errors, while the addition of the structure-aware factor $\phi$ further reduces HD95 and B0E in most configurations. Although some intermediate metrics are not strictly monotonic, introducing the source term $R$ consistently improves all metrics over the corresponding $\gamma+\phi$ configuration. The complete combination achieves the best performance across all evaluated datasets, backbones, and metrics, supporting the complementary roles of uncertainty-guided propagation, structure-aware interaction, and source-based stabilization.

\textbf{Parameter sensitivity.}
We further evaluated the sensitivity of UGCP to the number of update steps $T$, update step size $\theta$, temperature parameter $\tau$, and reference uncertainty level $u_0$ on ICA and ImageCAS with U-Net. One parameter was varied at a time while the remaining parameters were fixed at their default values. Because $\tau$ controls the sensitivity of the uncertainty-guided gates to local uncertainty differences, we evaluated it around the corresponding default values used for ICA and ImageCAS rather than enforcing an identical absolute range. DSC and clDice were reported to characterize regional overlap and centerline consistency, respectively.

\begin{figure}[t]
\centering
\includegraphics[width=\linewidth]{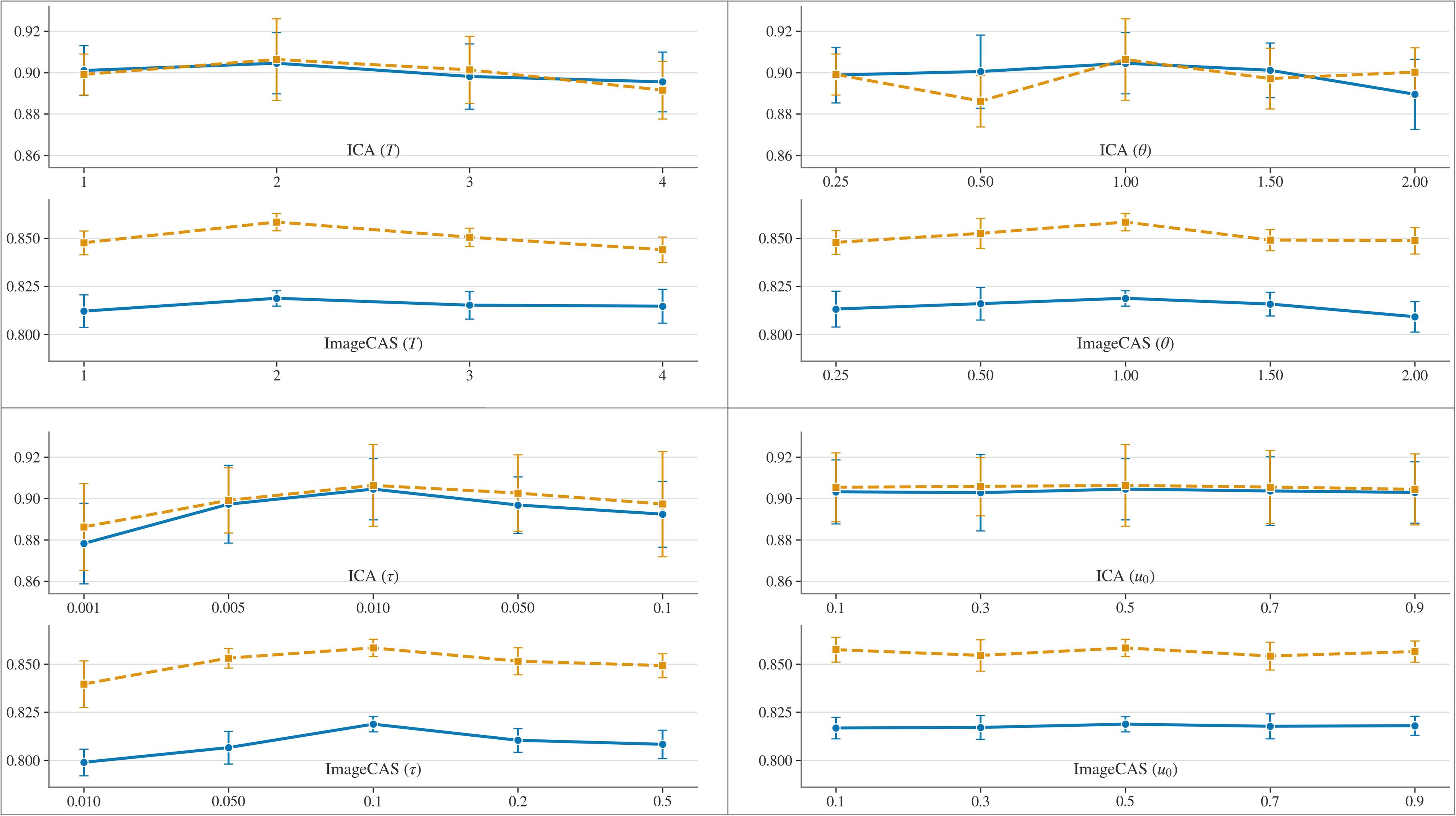} 
% =========================
% Parameter sensitivity
% =========================
\caption{
Parameter sensitivity of UGCP on ICA and ImageCAS using U-Net.
The evaluated parameters are the number of update steps $T$, update step size $\theta$, temperature parameter $\tau$, and reference uncertainty level $u_0$.
For each parameter, the upper and lower plots show the results on ICA and ImageCAS, respectively, while the remaining parameters are fixed at their default values.
}
\label{fig:parameter_sensitivity}
\end{figure}

As shown in Fig.~\ref{fig:parameter_sensitivity}, UGCP achieves the best overall performance at $T=2$ and $\theta=1$ on both ICA and ImageCAS, while further increasing either parameter generally degrades performance. The best temperature is $\tau=0.01$ on ICA and $\tau=0.1$ on ImageCAS. In contrast, performance varies only slightly across different values of $u_0$, indicating relatively low sensitivity to this parameter. These results support the default parameter settings adopted in our experiments.

 %%%%%%%%%%%%%%%%%%%%%%%%%%%%%%%%%%%%%%%%%%%%%%%%%%%%%%%%%%%%%%%%%%
%%        4.4   Comparison with CRF-Based Refinement       %%
%%%%%%%%%%%%%%%%%%%%%%%%%%%%%%%%%%%%%%%%%%%%%%%%%%%%%%%%%%%%%%%%%%
\subsection{Comparison with CRF-Based Refinement}

To compare UGCP with conventional inference-stage post-processing, we applied a standard non-trainable CRF to the baseline predictions on COSTA and FIVES using U-Net and SwinUNETR. The CRF was applied only during inference and did not modify the network architecture, training objective, or learned parameters. The network probability map was used to construct the unary term, while image intensities and spatial coordinates were used to define the pairwise terms. For COSTA, CRF refinement was performed independently on each 2D slice, whereas FIVES was refined at the full-image level. The CRF hyperparameters were fixed and applied consistently across all folds.

\begin{table}[!t]
\centering
\small
\renewcommand{\arraystretch}{0.90}
\setlength{\tabcolsep}{2pt}
\setlength{\aboverulesep}{1pt}
\setlength{\belowrulesep}{1pt}
\caption{
Comparison of UGCP with CRF post-processing on COSTA and FIVES using U-Net and SwinUNETR.
Best results for each dataset and backbone are shown in bold.
HD95 is reported in millimeters for COSTA and in pixels for FIVES.
}
\label{tab:crf_compatibility}

\begin{tabular*}{\linewidth}
{@{\extracolsep{\fill}} l l l c c c c @{}}
\toprule
Dataset
& Backbone
& Method
& DSC $\uparrow$
& clDice $\uparrow$
& HD95 $\downarrow$
& B0E $\downarrow$ \\
\midrule

\multirow{6}{*}{COSTA}
& \multirow{3}{*}{U-Net}
& Baseline
& $.8422 \pm .0043$
& $.8878 \pm .0033$
& $5.07 \pm 0.86$
& $123.25 \pm 5.37$ \\

&
& +CRF
& $.8435 \pm .0051$
& $.8992 \pm .0024$
& $5.01 \pm 0.72$
& $101.83 \pm 5.46$ \\

&
& +UGCP
& $\mathbf{.8542 \pm .0037}$
& $\mathbf{.9030 \pm .0044}$
& $\mathbf{3.34 \pm 0.53}$
& $\mathbf{67.60 \pm 4.35}$ \\

\cmidrule(lr){2-7}

& \multirow{3}{*}{SwinUNETR}
& Baseline
& $.8560 \pm .0025$
& $.8902 \pm .0058$
& $5.94 \pm 2.09$
& $105.90 \pm 9.62$ \\

&
& +CRF
& $.8609 \pm .0043$
& $.8961 \pm .0074$
& $5.43 \pm 1.88$
& $83.39 \pm 4.17$ \\

&
& +UGCP
& $\mathbf{.8790 \pm .0028}$
& $\mathbf{.9155 \pm .0037}$
& $\mathbf{2.09 \pm 0.43}$
& $\mathbf{43.92 \pm 2.43}$ \\

\midrule

\multirow{6}{*}{FIVES}
& \multirow{3}{*}{U-Net}
& Baseline
& $.8780 \pm .0089$
& $.8777 \pm .0110$
& $62.73 \pm 9.71$
& $120.90 \pm 10.50$ \\

&
& +CRF
& $.8801 \pm .0101$
& $.8795 \pm .0088$
& $57.14 \pm 9.26$
& $104.75 \pm 12.83$ \\

&
& +UGCP
& $\mathbf{.8831 \pm .0090}$
& $\mathbf{.8833 \pm .0109}$
& $\mathbf{50.70 \pm 9.98}$
& $\mathbf{89.10 \pm 6.20}$ \\

\cmidrule(lr){2-7}

& \multirow{3}{*}{SwinUNETR}
& Baseline
& $.8668 \pm .0090$
& $.8614 \pm .0108$
& $68.48 \pm 12.49$
& $161.98 \pm 6.74$ \\

&
& +CRF
& $.8693 \pm .0117$
& $.8623 \pm .0085$
& $65.91 \pm 8.24$
& $142.76 \pm 6.89$ \\

&
& +UGCP
& $\mathbf{.8711 \pm .0087}$
& $\mathbf{.8654 \pm .0103}$
& $\mathbf{58.08 \pm 11.48}$
& $\mathbf{130.66 \pm 10.50}$ \\

\bottomrule
\end{tabular*}

\end{table} 

As shown in Table~\ref{tab:crf_compatibility}, CRF improves the baseline across all evaluated metrics and settings, indicating that conventional post-processing can partially enhance vessel continuity and reduce local prediction errors. However, UGCP consistently achieves better results than CRF across all evaluated metrics and settings, with particularly pronounced improvements in HD95 and B0E. For example, on COSTA with SwinUNETR, UGCP reduces HD95 and B0E to $2.09$ mm and $43.92$, compared with $5.43$ mm and $83.39$ after CRF refinement. These results indicate that, under the evaluated settings, uncertainty-guided logit-space updates provide more effective structural refinement than the adopted CRF post-processing scheme.

 %%%%%%%%%%%%%%%%%%%%%%%%%%%%%%%%%%%%%%%%%%%%%%%%%%%%%%%%%%%%%%%%%%
%%        5   Discussion       %%
%%%%%%%%%%%%%%%%%%%%%%%%%%%%%%%%%%%%%%%%%%%%%%%%%%%%%%%%%%%%%%%%%%
\section{Discussion}
 %%%%%%%%%%%%%%%%%%%%%%%%%%%%%%%%%%%%%%%%%%%%%%%%%%%%%%%%%%%%%%%%%%
%%        5.1 Prediction-update Behavior and Failure Cases     %%
%%%%%%%%%%%%%%%%%%%%%%%%%%%%%%%%%%%%%%%%%%%%%%%%%%%%%%%%%%%%%%%%%%
\subsection{Analysis of Prediction Updates and Failure Cases}

\textbf{Interpretation of the main findings.}
Across the four datasets and different backbone networks, the improvements brought by UGCP are more consistently reflected in clDice, HD95, and B0E than in DSC. For some stronger or task-specific backbones, the gain in DSC is limited or even slightly negative, while the structural metrics continue to improve. This behavior suggests that UGCP primarily benefits structural properties of vessel predictions, including centerline consistency, boundary accuracy, and connected-component consistency. In particular, correcting thin branches, local discontinuities, or spurious connections often affects only a small fraction of foreground pixels or voxels and therefore produces a limited change in regional overlap, while substantially influencing centerline, boundary, and connectivity measures. This explains why regional overlap and vascular structural quality do not necessarily improve in parallel.

The effectiveness of UGCP across generic CNNs, Transformers, and vessel-specific architectures further suggests that its structured prediction-update mechanism is not tied to a particular backbone design. This architecture independence is consistent with the placement of UGCP after the segmentation head, where it operates directly on the predicted logit state rather than relying on a specific internal feature-extraction mechanism. From a different perspective, the ablation results clarify the role of uncertainty in the update process. Evidential uncertainty supervision alone does not produce consistent improvements, whereas incorporating the uncertainty signal into the UGCP update process leads to more consistent gains. This indicates that the main benefit arises from using uncertainty to regulate local state interactions rather than from evidential supervision itself. Moreover, the paired statistical analyses based on out-of-fold sample-level results show that most observed improvements are statistically significant.

\begin{figure}[!htbp]
\centering
\includegraphics[width=\linewidth]{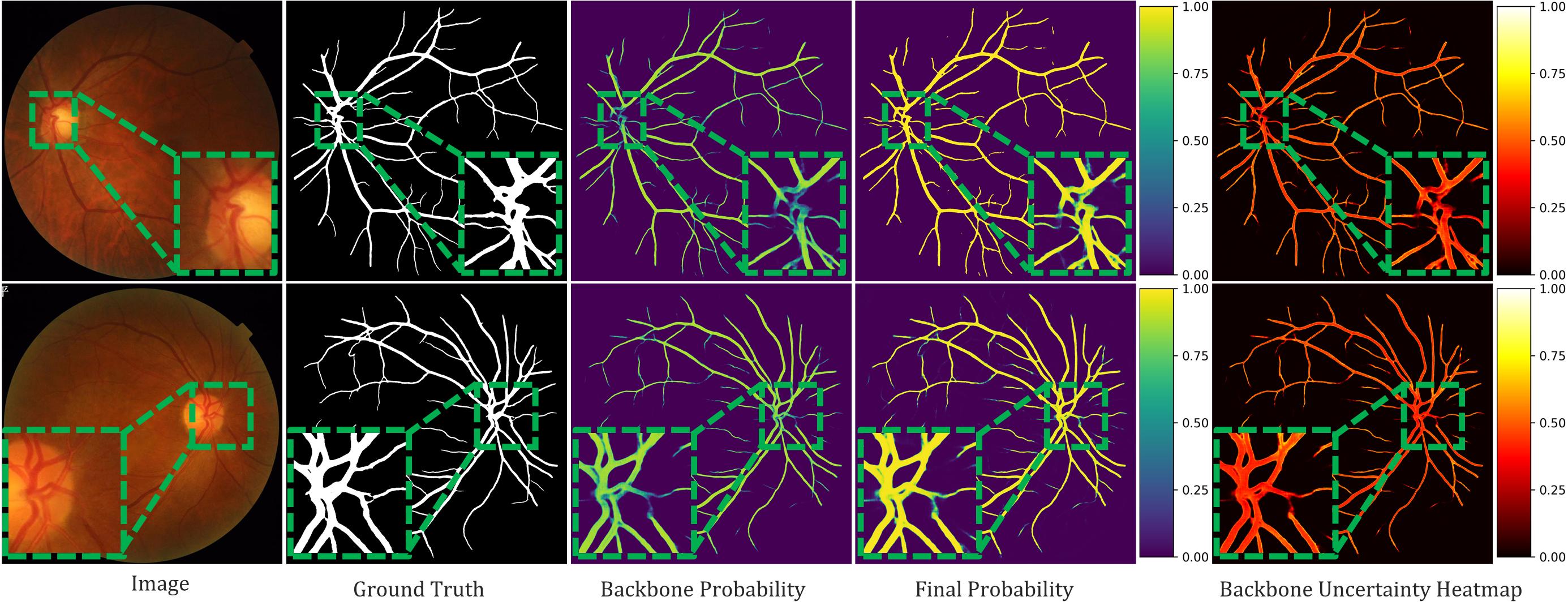}
% =========================
% 2D update visualization
% =========================
\caption{
Visualization of UGCP prediction-update behavior on representative FIVES examples.
From left to right: input image, ground truth, backbone probability, final probability, and backbone uncertainty.
}
\label{fig:heatmap_2d}
\end{figure}

\begin{figure}[!htbp]
\centering
\includegraphics[width=\linewidth]{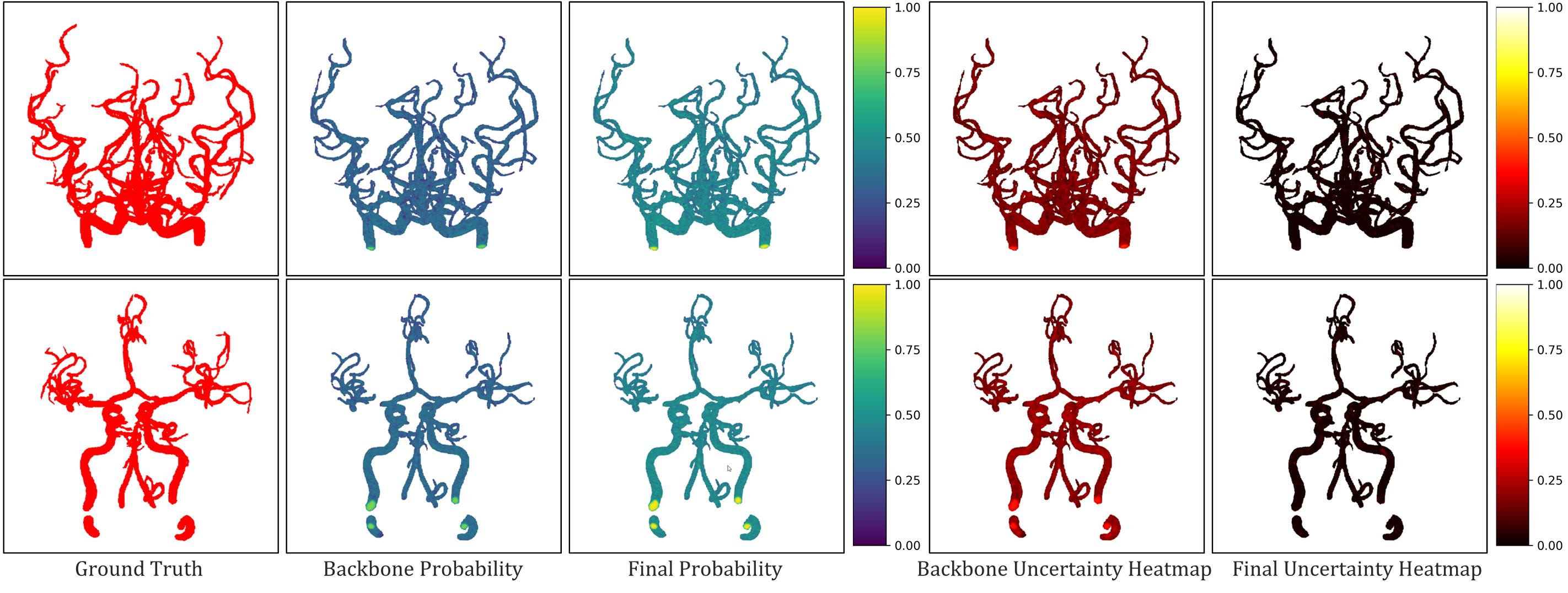}
% =========================
% 3D update visualization
% =========================
\caption{
Visualization of UGCP prediction-update behavior on representative COSTA examples.
From left to right: ground truth, backbone probability, final probability, backbone uncertainty, and final uncertainty.
}
\label{fig:heatmap_3d}
\end{figure}

\textbf{Analysis of prediction-update behavior.}
As shown in Figs.~\ref{fig:heatmap_2d} and \ref{fig:heatmap_3d}, weak or discontinuous vessel responses and ambiguous boundary regions in both the 2D and 3D examples are generally characterized by low vessel probabilities. The corresponding uncertainty maps provide complementary information about these unstable regions, serving as an additional signal for regulating local updates. After UGCP updates, the vessel probabilities in these regions increase, resulting in more continuous vessel responses. In contrast, high-confidence vessel trunks and background regions exhibit relatively limited changes. These observations suggest that UGCP does not uniformly smooth the entire probability map, but instead concentrates its updates on weak or unstable response regions where compatible neighboring evidence is available.

This selective update behavior is consistent with the design of UGCP. The directional gate regulates information flow according to the relative uncertainty between neighboring locations, the structure-aware modulation suppresses propagation that is incompatible with the local vascular structure, and the source term reduces the drift of reliable responses during updating. The parameter sensitivity analysis in Fig.~\ref{fig:parameter_sensitivity} further supports the use of a finite and moderate number of update steps. Too few updates may provide insufficient local structural coordination, whereas excessive updates or an overly large step size may repeatedly propagate local deviations and disturb otherwise stable responses. Therefore, moderate update depths and step sizes provide a balance between structural correction and state stability.

\begin{figure}[!t]
\centering
\includegraphics[width=\linewidth]{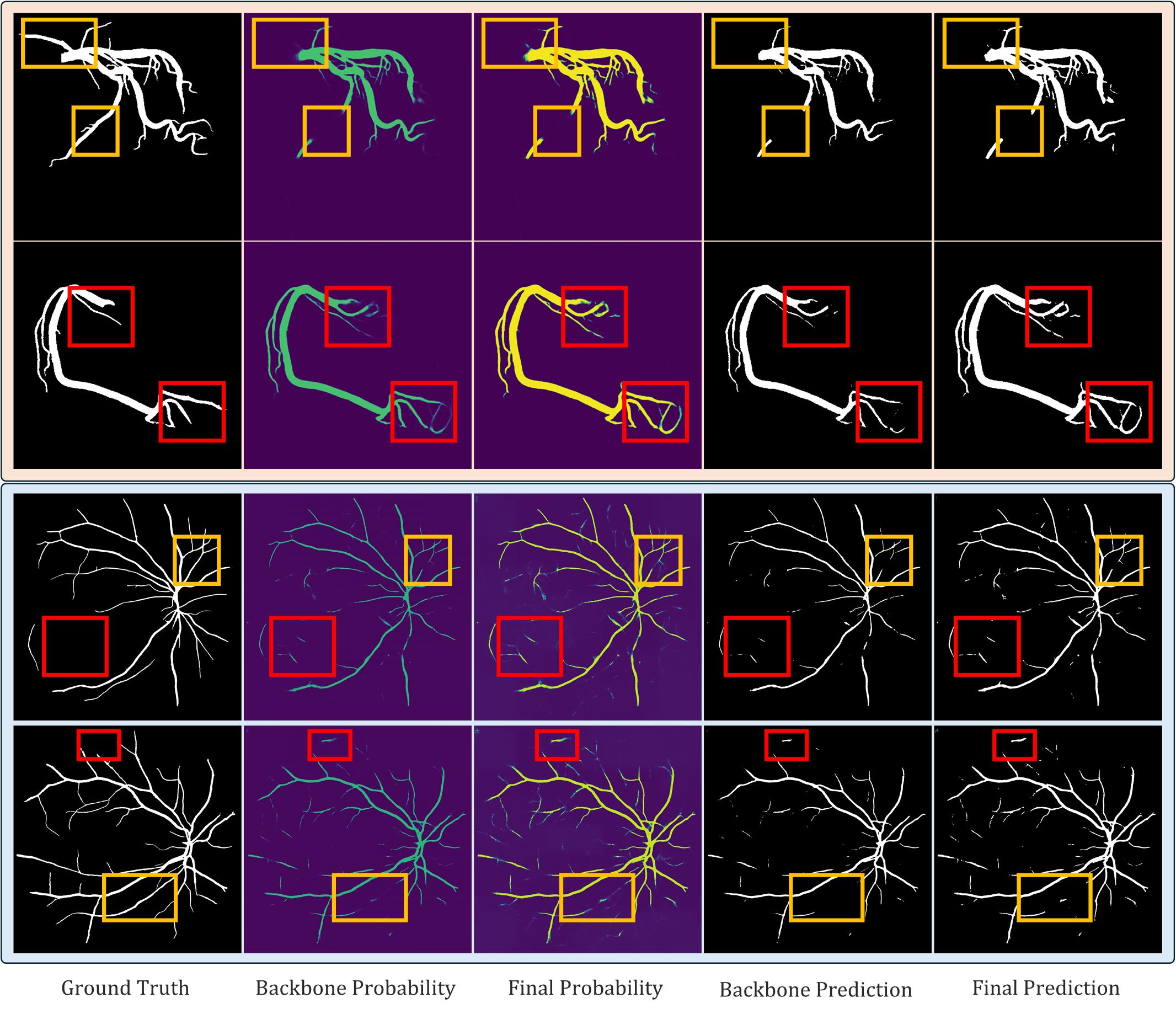}
% =========================
% Failure cases
% =========================
\caption{
Representative failure cases of UGCP on the 2D datasets.
Examples from ICA are shown in rows 1--2 and those from FIVES in rows 3--4.
The examples illustrate incomplete recovery of long vessel discontinuities and amplification of strong erroneous responses after UGCP.
}
\label{fig:failure}
\end{figure}

\textbf{Failure cases.}
As shown in Fig.~\ref{fig:failure}, UGCP remains dependent on the quality of the initial prediction and the availability of reliable local structural evidence. When a vessel discontinuity is too long to be bridged within the finite number of local update steps, UGCP can improve part of the missing response but may still fail to fully recover the vessel structure. Conversely, strong erroneous responses already present in the initial prediction can be further amplified during the update process. These cases indicate that UGCP is primarily effective for local refinement when sufficient structural evidence is retained in the initial prediction, rather than for recovering severe omissions or correcting strongly erroneous initial responses.

 %%%%%%%%%%%%%%%%%%%%%%%%%%%%%%%%%%%%%%%%%%%%%%%%%%%%%%%%%%%%%%%%%%
%%        5.2 Practical Implications and Computational Considerations      %%
%%%%%%%%%%%%%%%%%%%%%%%%%%%%%%%%%%%%%%%%%%%%%%%%%%%%%%%%%%%%%%%%%%
\subsection{Computational Efficiency and Practical Implications}

We evaluated the computational overhead of UGCP in terms of learnable parameters, GFLOPs, peak inference GPU memory, and inference time. All measurements were conducted with a batch size of $1$ on an NVIDIA RTX 4090 GPU with 24 GB of memory. The input sizes were $512 \times 512$ for the 2D models and $96 \times 96 \times 96$ for the 3D models. Peak inference GPU memory denotes the maximum GPU memory allocated during a single forward pass.

\begin{table}[!t]
\centering
\small
\setlength{\tabcolsep}{3pt}
\caption{
Computational complexity of the baseline and UGCP-enhanced models in the 2D and 3D settings.
Peak memory denotes the maximum GPU memory allocated during a single forward pass.
}
\label{tab:complexity}
\begin{tabular*}{\linewidth}{@{\extracolsep{\fill}} l c l c c c c @{}}
\toprule
Setting & Backbone & Method & Params (M) & GFLOPs & Memory (GB) & Time (ms) \\
\midrule

\multirow{6}{*}{2D}
& \multirow{2}{*}{U-Net}
& Baseline & $1.625$ & $9.520$ & $0.030$ & $2.272$ \\
&
& +UGCP & $1.625$ & $9.646$ & $0.122$ & $3.779$ \\

\cmidrule(lr){2-7}

& \multirow{2}{*}{SwinUNETR}
& Baseline & $6.302$ & $38.166$ & $0.231$ & $9.687$ \\
&
& +UGCP & $6.302$ & $38.292$ & $0.231$ & $10.840$ \\

\cmidrule(lr){2-7}

& \multirow{2}{*}{FR-UNet}
& Baseline & $7.375$ & $471.280$ & $0.876$ & $27.979$ \\
&
& +UGCP & $7.375$ & $471.436$ & $0.876$ & $29.123$ \\

\midrule

\multirow{6}{*}{3D}
& \multirow{2}{*}{U-Net}
& Baseline & $4.806$ & $22.991$ & $0.089$ & $3.547$ \\
&
& +UGCP & $4.807$ & $24.768$ & $0.455$ & $7.953$ \\

\cmidrule(lr){2-7}

& \multirow{2}{*}{SwinUNETR}
& Baseline & $15.703$ & $167.290$ & $1.259$ & $53.601$ \\
&
& +UGCP & $15.704$ & $169.067$ & $1.259$ & $58.220$ \\

\cmidrule(lr){2-7}

& \multirow{2}{*}{DSCNet}
& Baseline & $7.226$ & $443.926$ & $6.562$ & $411.806$ \\
&
& +UGCP & $7.226$ & $445.745$ & $6.611$ & $413.919$ \\

\bottomrule
\end{tabular*}
\end{table}
As shown in Table~\ref{tab:complexity}, UGCP introduces a negligible number of additional learnable parameters, while the increase in GFLOPs remains below approximately $7.7\%$ across all evaluated backbones. The additional inference latency ranges from approximately $1.1$ to $4.6$ ms. For the lightweight U-Net models, the relative increases in inference time and peak GPU memory are more noticeable because of their low baseline computational cost; nevertheless, the absolute peak memory remains below $0.5$ GB and the additional latency remains below $4.5$ ms. In contrast, for the computationally heavier SwinUNETR, FR-UNet, and DSCNet backbones, integrating UGCP introduces only small additional GFLOPs, memory consumption, and inference latency. Overall, UGCP introduces a modest absolute computational overhead, although its relative cost is more apparent for lightweight backbones.

Under this computational profile, the observed improvements in vascular structural quality could provide a more suitable segmentation basis for downstream analyses that depend on complete vascular geometry, including branch tracking, geometric quantification, stenosis analysis, and hemodynamic modeling. However, the present study focuses on segmentation performance and computational efficiency, and the downstream benefits of these structural improvements have not been directly evaluated. Further task-specific validation and clinical workflow studies are therefore required to determine whether the observed segmentation improvements translate into measurable downstream benefits.

\subsection{Limitations and Future Directions}

Despite the relatively consistent improvements across different datasets and backbone networks, UGCP remains a local refinement mechanism whose effectiveness depends on the quality of the initial prediction and the availability of reliable local structural evidence. Its use of fixed $4$-neighborhoods in 2D and $6$-neighborhoods in 3D, together with a finite number of update steps, limits the range over which structural information can be propagated. Consequently, severe omissions or long vessel discontinuities cannot always be fully recovered. In addition, the current formulation has only been validated for binary vessel segmentation. Future work will explore multi-scale or adaptive neighborhoods, graph-based interactions, and class-specific state representations to better model long-range structures and multi-class relationships.

Although this study covers multiple public datasets, imaging modalities, and backbone architectures, five-fold cross-validation cannot replace independent cross-institutional validation. The generalizability of UGCP across scanners, imaging protocols, and patient populations therefore requires further evaluation on independent multi-center datasets with appropriate expert annotation. Variations in annotation protocols and expert interpretation may also affect the reference vessel structures and topology-based evaluation, motivating future studies with multi-expert annotations.

In addition, the task-level and clinical utility of UGCP remains unverified. Dedicated downstream evaluations and clinical workflow studies are therefore required to determine whether the observed structural improvements translate into measurable practical benefits.

Another limitation is that the calibration and error-localization capability of the estimated uncertainty have not been systematically evaluated. Further uncertainty-specific analyses and reader studies are needed before the uncertainty estimates can be considered reliable for applications such as quality control or human review.

 %%%%%%%%%%%%%%%%%%%%%%%%%%%%%%%%%%%%%%%%%%%%%%%%%%%%%%%%%%%%%%%%%%
%%        6 Conclusion       %%
%%%%%%%%%%%%%%%%%%%%%%%%%%%%%%%%%%%%%%%%%%%%%%%%%%%%%%%%%%%%%%%%%%
\section{Conclusion}

In this work, we proposed UGCP, a differentiable and backbone-agnostic structured inference module for vessel segmentation. UGCP reformulates the initial logit prediction as a finite-step update process and regulates local state interactions through uncertainty-guided propagation, structure-aware modulation, and source-based stabilization. Across four public datasets and generic CNN and Transformer backbones as well as vessel-specific architectures, UGCP generally maintained or improved regional overlap while increasing clDice by $0.0032$--$0.0253$ and reducing B0E by approximately $10.9\%$--$60.1\%$. Ablation studies showed that the performance gains primarily arise from the proposed structured update mechanism rather than from evidential uncertainty supervision alone, while parameter sensitivity analysis and prediction-update visualizations further characterized the behavior of the proposed updates. UGCP also introduces negligible parameter growth and modest absolute computational overhead, supporting its use as a practical plug-in module for structurally consistent vessel segmentation. Future work will investigate multi-scale and adaptive propagation, multi-class vascular modeling, independent multi-center validation, and downstream clinical applications.

%%%%%%%%%%%%%%%%%%%%%%%%%%%%%%%%%%%%%%%%%%%%%%%%%%%%%%%%%%%%%%%%%%
%%        Declarations       %%
%%%%%%%%%%%%%%%%%%%%%%%%%%%%%%%%%%%%%%%%%%%%%%%%%%%%%%%%%%%%%%%%%%

\section*{Declarations}

\textbf{Declaration of competing interest.}

The authors declare no competing financial interests or personal relationships that could have influenced this work.

\textbf{Data availability.}

The FIVES, ICA, ImageCAS, and COSTA datasets are publicly available from their original sources cited in this paper.

\textbf{Funding.}

This work was partially supported by the American Heart Association under Award No. 25AIREA1377168 (Principal Investigator: Chen Zhao). 

\textbf{Declaration of generative AI and AI-assisted technologies in the writing process.}

During the preparation of this work, the authors used ChatGPT (OpenAI) to improve language readability, grammar, and code formatting. After using this tool, the authors reviewed and edited the content as needed and take full responsibility for the content of the published article.

\section*{CRediT authorship contribution statement}

Huan Huang: Methodology, Experiments, Implementation, Writing -- original draft.

Michele Esposito: Clinical interpretation, Discussion, Manuscript review.

Chen Zhao: Conceptualization, Supervision, Writing -- review and editing. 

\bibliographystyle{elsarticle-num}
\bibliography{reference}
\end{document}